%% file: main.tex
\documentclass[sigconf,edbt]{acmart-edbt2023}

\usepackage{subcaption}
\usepackage{xspace}
\usepackage{booktabs} 

\graphicspath{{./figs/}}

\newcommand{\stitle}[1]{\noindent\textbf{#1}}

\newcommand{\meanvar}{\textsc{MeanVar}\xspace}
\newcommand{\likely}{\textsc{SUL}\xspace}

\newcommand{\lar}{{\sffamily{LAR}}\xspace}
\newcommand{\crime}{{\sffamily{Crime}}\xspace}
\newcommand{\semisynth}{{\sffamily{SemiSynth}}\xspace}
\newcommand{\synth}{{\sffamily{Synth}}\xspace}

\setcopyright{rightsretained}

\acmDOI{}

\acmISBN{978-3-89318-092-9}

\acmConference[EDBT 2023]{26th International Conference on Extending Database Technology (EDBT)}{28th March-31st March, 2023}{Ioannina, Greece}
\acmYear{2023}

\begin{document}

\title{Auditing for Spatial Fairness}


\author{Dimitris Sacharidis}
\email{dimitris.sacharidis@ulb.be}
\affiliation{%
  \institution{Université Libre de Bruxelles}
  \city{Brussels}
  \country{Belgium}
}

\author{Giorgos Giannopoulos}
\email{giann@athenarc.gr}
\affiliation{
\institution{IMSI/Athena Research Center}
\city{Athens}
\country{Greece}
}

\author{George Papastefanatos}
\email{gpapas@athenarc.gr}
\affiliation{
\institution{IMSI/Athena Research Center}
\city{Athens}
\country{Greece}
}

\author{Kostas Stefanidis}
\email{konstantinos.stefanidis@tuni.fi}
\affiliation{
\institution{Tampere University}
\city{Tampere}
\country{Finland}
}

\renewcommand{\shortauthors}{D. Sacharidis et al.}


\begin{abstract}
This paper studies algorithmic fairness when the protected attribute is location. To handle protected attributes that are continuous, such as age or income, the standard approach is to discretize the domain into predefined groups, and compare algorithmic outcomes across groups. However, applying this idea to location raises concerns of gerrymandering and may introduce statistical bias. Prior work addresses these concerns but only for regularly spaced locations, while raising other issues, most notably its inability to discern regions that are likely to exhibit spatial unfairness. Similar to established notions of algorithmic fairness, we define spatial fairness as the statistical independence of outcomes from location. This translates into requiring that for each region of space, the distribution of outcomes is identical inside and outside the region. To allow for localized discrepancies in the distribution of outcomes, we compare how well two competing hypotheses explain the observed outcomes. The null hypothesis assumes spatial fairness, while the alternate allows different distributions inside and outside regions. Their goodness of fit is then assessed by a likelihood ratio test. If there is no significant difference in how well the two hypotheses explain the observed outcomes, we conclude that the algorithm is spatially fair.

\end{abstract}




\maketitle

\input{intro}

\input{related}

\input{problem}
\input{experiments}

\input{conclusion}


\bibliographystyle{ACM-Reference-Format}
\bibliography{biblio}

\clearpage
\input{appendix}

\end{document}

%% file: intro.tex
\section{Introduction}
\label{sec:intro}

Algorithmic fairness refers to the notion that the algorithm (e.g., the machine, an ML model, an AI system) should not discriminate against individuals. Typically, discrimination is defined over groups of people that are considered \emph{protected}, such as race or gender minorities. Abstractly, fairness requires that each population group, determined by a specific value to the \emph{protected attribute} (e.g., race, gender), is on average treated or affected by the algorithm in the same manner.
To make this requirement concrete, one first needs to define a \emph{measure} to appropriately quantify the behavior, performance, etc.\ of the algorithm. 
Then, algorithmic fairness is achieved when the measure is statistically \emph{independent} of the protected attribute.
In practice, this mandates that the measure is distributed equally among protected groups.
Fairness notions differ in how they define the measure. For example, statistical parity considers the positive rate (how often the algorithm assigns the positive/desirable class) as the measure, whereas equal opportunity \cite{2016_NIPS_HPS} considers the true positive rate (how often the algorithm correctly assigns the positive/desirable class).

\stitle{Motivation.}
In many cases, it is important to ensure that the algorithm does not discriminate against individuals on the basis of their location (place of origin, home address, etc.). That is, we consider location as the protected attribute and we want the algorithm to exhibit \emph{spatial fairness}. For example, consider an algorithm that decides whether mortgage loan applications are accepted or not.\footnote{In practice, an algorithm would compute a ``credit score'', based on which a human would take the decision. In that case, we would like these scores to be spatially fair.} 
It is desirable that its decisions do not discriminate on the basis of the home address of the applicant. This could be to avoid \emph{redlining}, i.e., indirectly discriminating based on ethnicity/race due to strong correlations between the home address and certain ethnic/racial groups, or to avoid \emph{gentrification}, e.g., when applications in a poor urban area are systematically rejected to attract wealthier people. As another example, consider crime forecasting, where an algorithm predicts how likely a crime is to occur in a particular area. It is desirable that the algorithm is spatially fair in terms of its accuracy. That is, we require the predicted crime rate to not differ greatly than the observed crime rate \emph{in all areas}. This could be to avoid \emph{under-} and \emph{over-policing}, and the sense of injustice they are typically associated with.

In cases like these, there is the common need for a principled method to \emph{audit} an algorithm for spatial fairness, i.e., examine its outcomes and answer the question ``\emph{is it fair?}''. More importantly, if this answer is negative, the method should credibly \emph{testify}, i.e., provide as evidence a region that is most likely to suffer from discrimination, and thus answer the question ``\emph{where is it unfair?}''.

\stitle{Challenges.}
Unlike typical protected attributes, such as race and gender, location is a continuous attribute.
The group-based definition of fairness does not apply, in a straightforward manner, to \emph{continuous protected attributes}. Rather, the standard approach is to first \emph{discretize} the continuous domain to create groups, e.g., age or income groups, and then compare the outcomes for each group.
The same idea can be applied to location, by defining non-overlapping spatial partitions (e.g., city blocks, zipcodes, districts), and computing the measure in each. This leads to a partitioning-based definition, according to which an algorithm is spatially fair if the measures per partition are equal.\footnote{A note on terminology: a partitioning of the space consists of a set of non-overlapping regions, called partitions, that collectively cover the space.}

However, this simple partitioning-based definition has two drawbacks. Location is highly susceptible to \emph{gerrymandering} \cite{pmlr-v80-kearns18a,2022_AAAI_XHJ+}, which is the act of purposefully defining a partitioning of the space (via setting the partition boundaries, extents, shapes) so that the partition measures appear non-discriminatory. Moreover, conclusions drawn from comparing spatial aggregates highly depend on the shape and scale of the partitions (the areal units), a well-known source of statistical bias termed the \emph{modifiable areal unit problem} (MAUP) \cite{2022_AAAI_XHJ+}.

A better partitioning-based measure of spatial fairness that tries to address these two drawbacks is proposed in \cite{2022_AAAI_XHJ+}. Briefly, the idea is to superimpose a high-resolution grid over the space, and consider all possible rectangular, grid-aligned partitionings of the space. In each partitioning, the variance of the measure in the partitions is computed. Then, the mean variance across all partitionings, hereafter denoted as \meanvar, is computed. Lower values of \meanvar suggest lower variance across the partitions in all partitionings and hence more fairness.

\begin{figure}
\centering
\begin{subfigure}[b]{0.48\columnwidth}
\includegraphics[width=\linewidth]{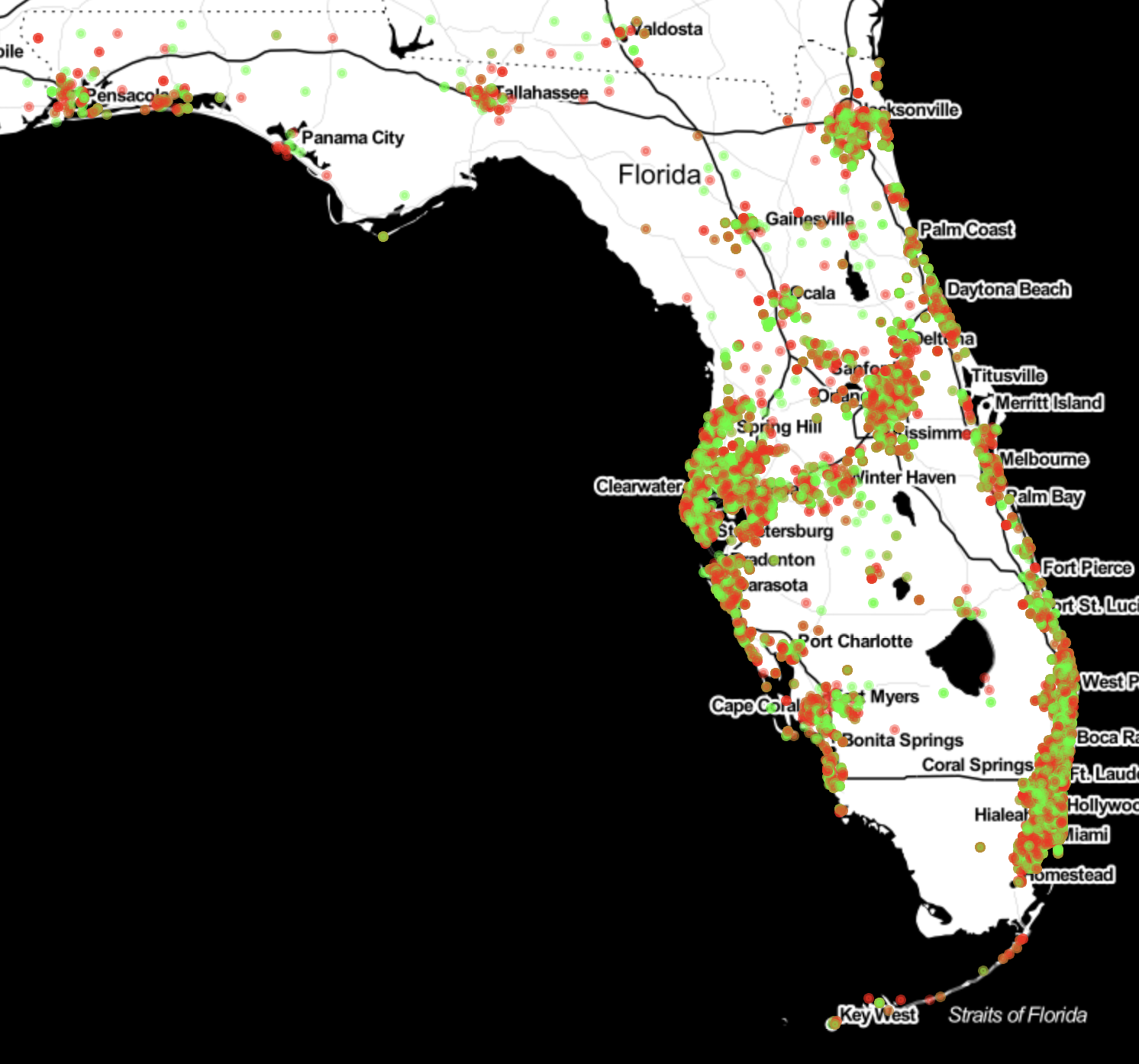}
\caption{Spatially fair-by-design outcomes distributed in Florida; each outcome has $0.5$ probability of being positive.}
\end{subfigure}%
\hspace{0.04\columnwidth}%
\begin{subfigure}[b]{0.48\columnwidth}
\includegraphics[width=\linewidth]{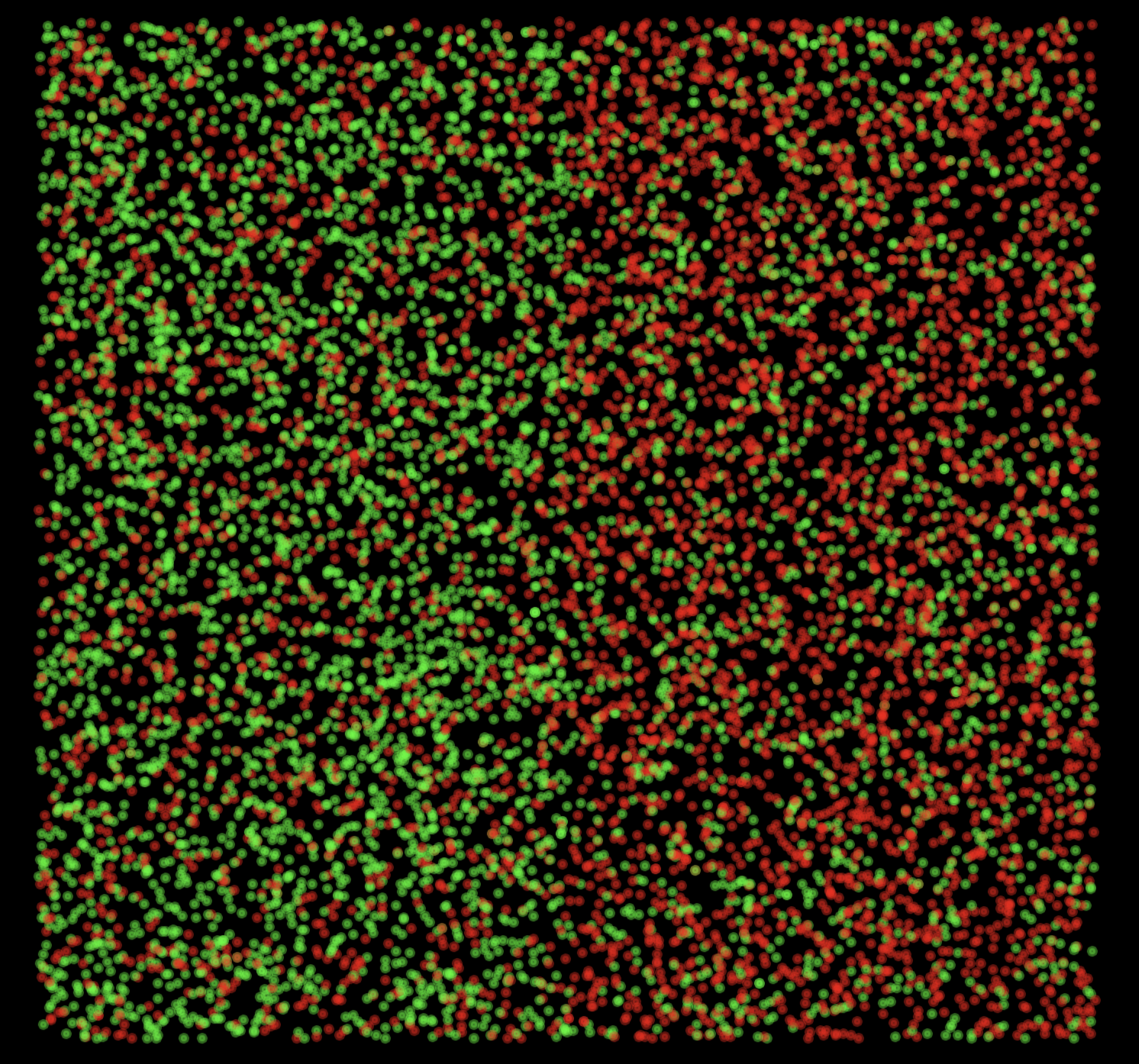}
\caption{Spatially unfair-by-design outcomes uniformly distributed; left half has twice more positive outcomes than right half.}
\end{subfigure}
\caption{Two spatial distributions each with $n=10,000$ outcomes among which $p=5,000$ are positive. According to \cite{2022_AAAI_XHJ+}, the fair-by-design distribution is less fair than the unfair-by-design distribution (\meanvar of $0.0522$ vs.\ $0.0431$).}
\label{fig:audit_fair}
\end{figure}


Note that \meanvar is designed to assess the spatial fairness of outcomes that are \emph{regularly distributed} in space: in each cell of the superimposed grid (or equivalently in each partition of the partitioning with the highest resolution) there is roughly the same number of outcomes. As a consequence, computing \meanvar in the general case of outcomes arbitrarily distributed in space leads to counter-intuitive results. Concretely, \emph{\meanvar cannot reliably audit an algorithm for spatial fairness}. To illustrate this, consider the example presented in Figure~\ref{fig:audit_fair}.
The algorithm producing the \semisynth dataset depicted on the left is \emph{spatially fair by design} as positive/desirable (resp. negative/undesirable) outcomes are randomly assigned to each location with a probability of $0.5$, and are indicated as green (resp. red) points.
The algorithm producing the \synth dataset depicted on the right is \emph{spatially unfair by design}, as the left half of the area contains twice as many positive outcomes as the right half does. However, if we assess spatial fairness according to \cite{2022_AAAI_XHJ+}, we find that the unfair-by-design algorithm has a lower \meanvar of $0.0431$ and is thus considered more fair than the fair-by-design algorithm with a higher \meanvar of $0.0522$. This occurs because the spatial distribution of outcomes in Figure~\ref{fig:audit_fair}(a) is non-regular, and thus the number of outcomes within partitions varies greatly. Despite the fact that the ratio of positives is on average the same in each partition (by design), there exist several sparse partitions with few outcomes that are predominantly positive or negative, and which increase the variance of the measure. The example demonstrates that it is impossible to set a threshold for \meanvar such that it distinguishes a fair from an unfair algorithm. Therefore, \meanvar cannot answer the ``\emph{is it fair?}'' question.


\begin{figure}
\centering
\begin{subfigure}[b]{0.48\columnwidth}
\includegraphics[width=\linewidth]{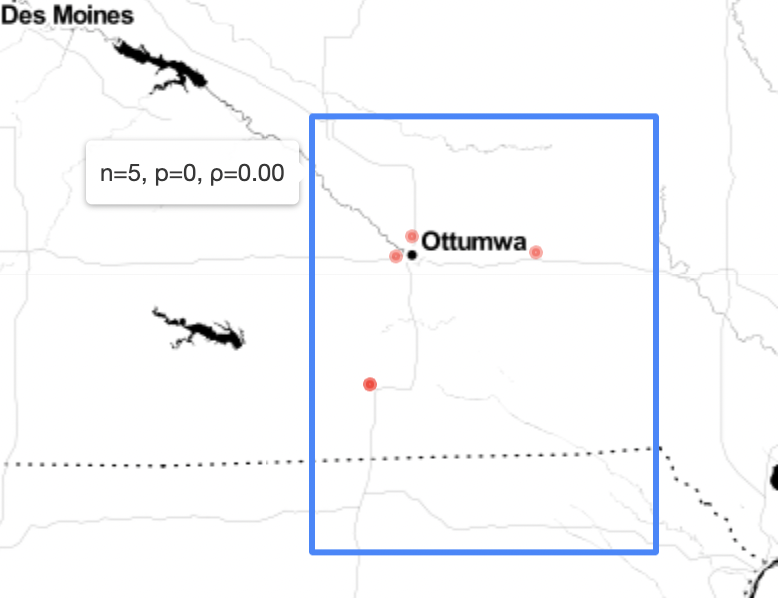}
\caption{A suspicious region in Iowa that makes the largest contribution to \meanvar.}
\end{subfigure}%
\hspace{0.04\columnwidth}%
\begin{subfigure}[b]{0.48\columnwidth}
\includegraphics[width=\linewidth]{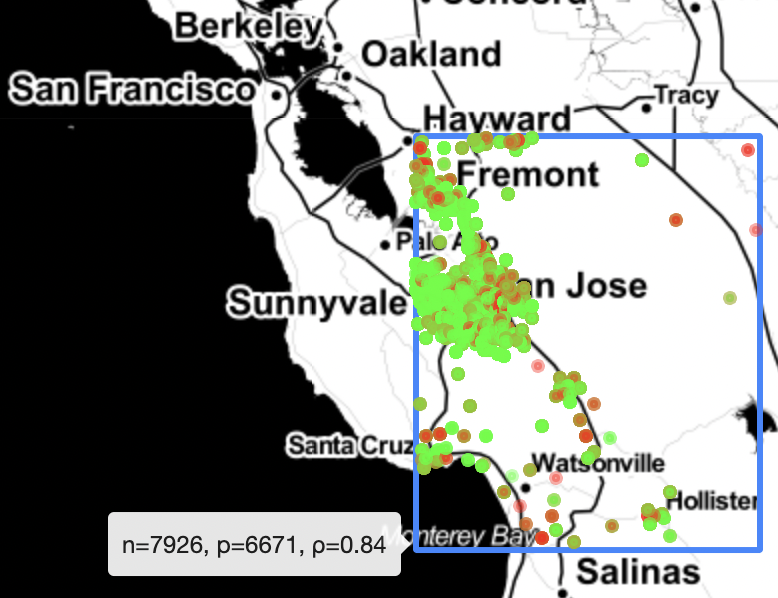}
\caption{A region in California that testifies for spatial unfairness according to our framework (p-value $<0.005$).}
\end{subfigure}
\caption{Regions most likely to be unfair according to different spatial fairness definitions for the \lar dataset described in Section~\ref{sec:exp:data}. Depicted are the the number of observations ($n$), the number of positive outcomes ($p$), and the local positive rate ($\rho$); the overall positive rate is $0.62$.}
\label{fig:pos_rates}
\end{figure}

Even if \meanvar cannot discern fairness, one may wonder if it can be utilized to identify \emph{suspicious regions}, i.e., regions that are likely to exhibit discrimination. The obvious way to search for suspicious regions is to consider the partitions that contribute the most to the \meanvar value. 
These are the partitions whose measures are the farthest away from the partitioning mean and thus take extreme values. For the reasons discussed before, these partitions are likely to be sparse and predominantly positive or negative. An example is given in Figure~\ref{fig:pos_rates}(a), where the depicted partition with just five negative outcomes ties for the largest contribution to \meanvar. On a first look, this appears to suggest an area of discrimination. However, this result is not statistically significant: it is not that uncommon to find a region that contains at least five negatives and no positives by chance (see Figure~\ref{fig:random}). Therefore, even though \meanvar can identify regions with extreme measures, arguably it cannot meaningfully answer the ``\emph{where is it unfair?}'' question.


\stitle{Our Solution.}
In this work, we propose a practical definition for spatial fairness that enables \emph{auditing} and \emph{testifying}, i.e., that can answer both important questions regarding the algorithm, ``\emph{is it fair?}'' and if not ``\emph{where is it unfair?}''.
The definition applies to the most general case and does not assume regularly distributed observations.

Recall that algorithmic fairness is when some measure quantifying the behavior or performance of the algorithm is statistically independent of the protected attribute. In our setting, location is the protected attribute, so we naturally consider an algorithm to be \emph{spatially fair}, if the measure is \emph{independent of location}.
This implies that for any region of the space, the distribution of the measure \emph{inside} and \emph{outside} the region should be the same.

To operationalize this definition, there are several challenges.
The most complicating is how to determine the distribution of the measure within a region. If the region covers many observations, the observed (empirical) distribution of the measure is a good proxy for the actual distribution. Otherwise, what we observe in a sparse or small region might differ dramatically from what we observe outside it. Note that this issue does not manifest itself in categorical protected attributes (e.g., gender), simply because the number of observations per protected group (e.g., number of women and men) are typically very large.

To address this issue, instead of looking at the observed distribution, we want to express \emph{how likely} it is to observe such a distribution if the algorithm was fair. Intuitively, we should expect to find a region with only four negative points clustered together even when the algorithm is fair. Conversely, we should not expect to find a region that contains thirty negative points alone --- observing such a region should be a stronger indication that the algorithm is not fair.

Inspired by the work in spatial scan statistics \cite{1997_CSTM_K,2010_SM_JKR,2020_SIGSPATIAL_XBLS}, we form two hypotheses and seek to quantify which one is a better fit for the data observed. The \emph{null hypothesis} is that of spatial fairness, i.e., there is a single distribution that controls how the measure is distributed in the space, or mnemonically \emph{inside} = \emph{outside}. The \emph{alternate hypothesis} states that there is a difference in the distribution inside and outside a region, or \emph{inside} $\neq$ \emph{outside}. Given the observed data, we can determine the maximum likelihood for each hypothesis, compute their ratio, and test whether this likelihood ratio is statistically significant at a desired level.

The remainder of this paper is organized as follows. Section~\ref{sec:related} reviews important concepts from algorithmic fairness and scan statistics. Section~\ref{sec:approach} introduces our definition of fairness and the auditing framework. Section~\ref{sec:exp} presents some results of our framework on several datasets. Section~\ref{sec:concl} concludes this work.

%% file: related.tex
\section{Related Work}
\label{sec:related}

\subsection{Algorithmic Fairness}

Algorithmic fairness has seen various definitions \cite{DBLP:journals/vldb/PitouraSK22}. One important distinction \cite{DHP+12} is between \textit{individual fairness} definitions are based on the premise that similar entities should be treated similarly, and \textit{group fairness} definitions group entities based on the value of one or more protected attributes and ask that all groups are treated similarly. In this work, we consider the definition of the latter category. 
Note that fairness by unawareness, where the protected attribute is not considered by the algorithm is not sufficient in many cases, due to the presence of other attributes that might be correlated with them.

Let $X$ denote a set of features that describe an individual, and let $A$ denote the protected attribute(s); for ease of presentation assume that $A$ is a binary valued attribute, where $A=1$ indicates the protected group. 
For what follows we assume that the algorithm is a binary classification model. 
Therefore, let $Y$ denote the actual class of the individual (the ground truth), with the positive class $Y=1$ denoting the desirable/favorable class. For example, the positive class might correspond to the acceptance of a loan application, the hiring of a candidate, the granting of parole, etc. Moreover, let $\hat{Y}$ denote the predicted output of the algorithm; note that although we do not explicitly show it, $\hat{Y}$ depends on $X$.

Statistical approaches to group fairness can be distinguished as \cite{FSV+19}: \textit{base rates approaches} that use only the output $\hat{Y}$ of the algorithm, and \textit{accuracy approaches} that use both the output $\hat{Y}$ of the algorithm and the ground truth $Y$. Base rate fairness compares the probability $P(\hat{Y}=1|A=1)$ that one individual receives the favorable outcome when they belong to the protected group with the corresponding probability $P(\hat{Y}=1|A=0)$ that one receives the favorable outcome when they belong to the non-protected group. To compare the two, we may take their ratio \cite{FFM+15} or their difference \cite{CV10}. 
When the probabilities of a favorable outcome are equal for the two groups, we have a special type of fairness termed \textit{statistical parity}. Statistical parity preserves the input ratio, that is, the demographics of the individuals receiving a favorable outcome are the same as the demographics of the underlying population. Statistical parity is a natural way to model equity: members of each group have the same chance of receiving the favorable output.

Accuracy-based fairness warrants that various types of classification errors (e.g., true positives, false positives) are equal across groups. Depending on the type of classification errors considered, the achieved type of fairness takes different names \cite{2016_NIPS_HPS}. For example, the case in which we ask that $P(\hat{Y}=1| Y = 1, A=1)$ = $P(\hat{Y}=1|Y =1, A=0)$ (i.e., the case of equal true positive rate for the two groups) is called \textit{equal opportunity}. Similarly, the case in which both the true positive rate and the false positive rate are equal for the two groups, is called \textit{equal odds}.

These notions can be generalized under a common idea. Let $M$ indicate the event/resource that is to be distributed fairly among individuals. In the case of statistical parity, the resource is positive rate and the event is $M = \hat{Y}$. In equal opportunity, the resource is true positive rate and thus the event is $M = \hat{Y}|Y=1$. Equal odds considers the true positive rate and the false positive rate, which is modeled as $M = \hat{Y}|Y=0$. Then these fairness notion can be expressed as the requirement that $P(M|A=1) = P(M|A=0)$, or equivalently that $M$ is independent of $A$, i.e., $P(M|A) = P(M)$.

More recently, there have been approaches that define fairness from a causal perspective \cite{KusnerLRS17,NabiS18,KilbertusRPHJS17}. For example, we might require that $A$ does not \emph{cause} $M$, rather than simply requiring that $M$ is independent of $A$.

\subsection{Spatial Fairness}

The problem of spatial fairness has received little attention. As location is a continuous attribute, the standard approach is to discretize locations and apply the standard definitions \cite{WeydemannSW19}. This discretization approach may be susceptible to gerrymandering. To avoid this, \cite{2022_AAAI_XHJ+} considers all possible grid-based rectangular partitionings of the space into regions. Given a partitioning, the algorithm is considered fair if it exhibits roughly the same performance in each region. This is quantified as the variance of some performance metric $M$ across partitions. The algorithm is (perfectly) fair with respect to a partitioning if there is zero variance. The algorithm is (perfectly) fair if it is fair with respect to every partitioning, among a predefined set. 
In essence, this definition takes a brute-force approach to the problem: it applies the group-based approach, which is susceptible to gerrymandering, to each possible partition to eliminate the likelihood of gerrymandering.

\cite{2022_AAAI_XHJ+} defines an unfairness measure as the mean variance across partitionings. This measure is then used as an additional optimization goal in a learning process to produce classification models that do not exhibit spatial unfairness in this sense. As discussed in the previous section, and as we illustrate in Section~\ref{sec:exp}, the definition in \cite{2022_AAAI_XHJ+} has several shortcomings making it less meaningful in real-life applications where the observations are not regularly distributed in space.

\subsection{Scan Statistics}

Our notion of spatial fairness is inspired by the work in \emph{spatial scan statistics}. For a 1d domain like time, scan statistics answer questions such as whether there exists an unusually high concentration of events in some time period. In the spatial case, scan statistics answer similar questions, e.g., is there an unusually high concentration of a particular virus variant in some spatial area. Spatial scan statistics have been proposed for various underlying spatial processes, such as Bernoulli and Poisson \cite{1997_CSTM_K}, and multinomial distributions \cite{2010_SM_JKR}. Our work is directly related to the former type. Moreover, there is another line of work on detecting \emph{mixture areas} \cite{2020_SIGSPATIAL_XBLS, 2021_SIGSPATIAL_SSP}, e.g., areas that contain a low or high diversity of point categories.

%% file: problem.tex
\section{A Spatial Fairness Framework}
\label{sec:approach}

We consider binary classification tasks concerning individuals. Let $X$ denote a set of features that describe an individual. Among them, location $L$ is considered the protected attribute. 
Let $R$ denote a spatial region, and let $n(R)$ denote the number of individuals whose location falls in the region.
Moreover, let $Y$ denote the true class of the individual, with the positive class $Y=1$ denoting the desirable class.
Further, consider a classification model $\hat{Y}$ that takes as input the features $X$ of an individual and predicts their class $\hat{Y}(X)$. A summary of the important notation is in Table~\ref{tab:notation}.

In what follows, we define spatial fairness in terms of statistical parity, i.e., we quantify the positive rate of the model; the definitions can be adapted to the cases of equal opportunity and equal odds by replacing $\hat{Y}$ with $\hat{Y}|Y=1$ to account for the true positive and $\hat{Y}|Y=0$ for the false positive rates.

Let $\rho = Pr(\hat{Y}=1)$ denote the \emph{positive rate} ($pr$) of the model, i.e., the probability that the models predicts the positive class.
Restricting focus to a particular spatial region $R$, we define the \emph{local positive rate} as $\rho(R) = Pr(\hat{Y}=1 | L \in R)$, which is the $pr$ taking into account only individuals within the region $R$.

We now formulate an \emph{idealized} notion of spatial statistical parity,
by requiring that every local positive rate is equal to the positive rate of the model, i.e., $\rho(R) = \rho$ for all $R$; equivalently, all local positive rates should be equal.
This definition is impractical as it can only be satisfied by trivial classifiers that predict either 0 or 1. Thus, we must relax the definition to allow the local positive rates to differ from each other and the overall $pr$. The question is to what extent they can differ before we declare the model to be spatially unfair. To answer this, we define a statistical test, based on the Bernoulli spatial scan statistic \cite{1997_CSTM_K}, that decides whether the positive rate is \emph{homogeneous} across the space.

First, observe that we can interpret a positive rate $\rho$ as the probability that the model assigns an individual to the positive class; i.e., a Bernoulli trial with success probability $\rho$. Now, consider the group of people within a region $R$. The number of positive labels the model assigns follows the Binomial distribution with $n(R)$ trials and success probability $\rho$, denoted as $p(R) \sim B(n(R), \rho)$.

We now define two hypotheses and use a statistical test to determine which explains the observed data better.
The discussion that follows is based on the multinomial spatial scan statistic; an important difference is that we do not care for the direction of change of the statistic inside and outside a region.
The null hypothesis $\mathcal{H}_0$ states that in every region $R$ the number of positive labels, $p(R)$, follows the Binomial distribution $B(n(R), \rho_0)$, where $R, \rho_0$ are the parameters of the null hypothesis. The alternative hypothesis $\mathcal{H}_1$ states that there is a region $R$ such that $p(R)$ follows $B(n(R), \rho_0)$, while the number of positive labels outside $R$ follows a Binomial with a different success probability $\rho_1 \neq \rho_0$, i.e., $B(N-n(R), \rho_1)$. The alternate hypothesis has three parameters $R, \rho_0, \rho_1$.

To quantify which hypothesis explains the observed data better, we derive their maximum likelihoods, and compute the likelihood ratio.
Let us first consider the null hypothesis. The likelihood of $\mathcal{H}_0$ is given by $L_0(R, \rho_0) = \rho_0^{p(R)} (1-\rho_0)^{n(R)-p(R)}$, where $n(R)$ (resp. $p(R)$) denotes the number of individuals (resp. with positive labels) in a region $R$. For any region $R$, observe that the maximum likelihood is when $\rho_0 = p(R)/n(R)$. Across all regions, the likelihood takes the maximum value when $R$ is the entire space.
Therefore, the maximum likelihood value of the null hypothesis is $L^{max}_0 = \rho^{P} (1-\rho)^{N-P}$.

\begin{table}
\caption{Important Notation}
\label{tab:notation}
\small
\begin{tabular}{lp{5.8cm}}
\toprule
\textbf{Symbol} & \textbf{Meaning}\\
\midrule
$X$ & features of an individual\\
$L$ & location of an individual\\
$Y$ & actual outcome/class\\
$\hat{Y}$ & model prediction\\
$N$ & number of individuals\\
$P$ & num.\ of indiv.\ predicted to be in the positive class\\
$\rho = \frac{N}{P}$ & model's positive rate\\
$R$ & a spatial region\\
$n(R)$ & number of individuals in $R$\\
$p(R)$ & num.\ of indiv.\ in $R$ predicted to be in the pos.\ class\\
$\rho(R) = \frac{n(R)}{p(R)}$ & model's local positive rate in $R$\\
\bottomrule
\end{tabular}
\end{table}

Consider now the alternative hypothesis. Its likelihood is the product of two binomials, one for inside and one for outside the region $R$, each having a different success probability:
\begin{equation*}
\resizebox{1\hsize}{!}{
$L_1(R, \rho_0, \rho_1) = \rho_0^{p(R)} (1-\rho_0)^{n(R)-p(R)} \rho_1^{P-p(R)} (1-\rho_1)^{N-n(R)- (P-p(R))}$
}
\end{equation*}
First, we maximize the likelihood for a given $R$. Recall that $\mathcal{H}_1$ requires that $\rho_0 \neq \rho_1$. If the observed positive rate inside the region, $p(R)/n(R)$, is different from that outside the region, $(P-p(R))/(N-n(R))$, then the likelihood takes its maximum value when $\rho_0$ and $\rho_1$ take the values of the observed positive rates inside and outside the region, respectively. Otherwise, it cannot exceed the likelihood when $\rho_0=\rho_1 = \rho$. Concretely, we have:
\begin{equation}
L^{max}_1(R) =
\begin{cases}
L_1(R, \frac{p(R)}{n(R)}, \frac{P-p(R)}{N-n(R)}) & \ \text{if } \frac{p(R)}{n(R)} \neq \frac{P-p(R)}{N-n(R)}, \\
L^{max}_0 & \text{otherwise.}
\end{cases}
\label{eq:likelihood_value}
\end{equation}

We refer to the value in Eq.~\ref{eq:likelihood_value} as the \emph{spatial unfairness likelihood}, and denote it as \likely.
The next step is to identify the region that maximizes Eq.~\ref{eq:likelihood_value}. This is achieved by going over a predetermined set of regions $\mathcal{R}$. Let $R^* \in \mathcal{R}$ be the region that maximizes Eq.~\ref{eq:likelihood_value}. We compute the likelihood ratio test statistic of the alternative over the null hypothesis as $\tau = \frac{L^{max}_1(R^*)}{L^{max}_0}$. Note, that in practice, we compute and determine the difference of log-likelihoods.

To determine how significant the test statistic $\tau$ value is, we need to perform a Monte Carlo simulation to determine its distribution, as also performed in \cite{1997_CSTM_K,2020_SIGSPATIAL_XBLS}. Specifically, the simulation goes as follows.
We create alternate worlds assuming that the $N$ individuals are located as in our data, but their label is determined by a Bernoulli trial with success probability $\rho$. This mirrors the process generating the examples in Figure~\ref{fig:random}. For each alternate world, we compute the $\tau$ statistic. Suppose we simulate $w-1$ worlds, and the $\tau$ statistic of the real world ranks at the $k$-th highest position among all worlds. Then, the $p$-value of the real world's statistic is $k/w$.

We declare an algorithm as \emph{spatially fair} if the $p$-value of the aforementioned likelihood ratio test statistic is below some predefined significance level $\alpha$. In this case, the observed data are just as likely to be generated under the spatial fairness null hypothesis as under the alternate hypothesis. Otherwise, we have to reject the spatial fairness hypothesis. In this case, we may offer evidence that support the hypothesis that the algorithm is unfair. Providing evidence is the \emph{identification} process. We consider all examined regions that have a statistically significant likelihood ratio, and we rank them in decreasing order of their likelihood ratio. We then return the top-$k$ regions as evidence.

The computational complexity of our auditing framework is $O(M\cdot N \cdot Q)$, where $N$ is the number of regions to scan, $Q$ the average cost of a spatial range-count query, and $M-1$ is the number of Monte Carlo simulations.

%% file: experiments.tex
\section{Experiments}
\label{sec:exp}

Our experimental study investigates two research questions.
R1. In a setting compatible with prior work \cite{2022_AAAI_XHJ+}, does our spatial fairness definition present more meaningful results compared to prior work?
R2. Does our method identify regions that can be considered as potentially exhibiting spatial unfairness? Section~\ref{sec:exp:data} describes the dataset used, while Sections~\ref{sec:exp:partitioning} and \ref{sec:exp:unrestricted} answer the two research questions.

The code and datasets are available in GitHub.\footnote{\url{https://github.com/dsachar/AuditSpatialFairness}}

\subsection{Datasets}
\label{sec:exp:data}

In our experiments, we use two real, one semi-synthetic, and one synthetic datasets.
The first real dataset concerns about mortgage loan applications. Thanks to the Home Mortgage Disclosure Act, financial institutions in the US are required to publicly disclose information about mortgages. We download the modified loan/application register (LAR) records\footnote{\url{https://ffiec.cfpb.gov/data-publication/modified-lar/2021}} for Bank of America for the year 2021. The dataset contains information about over 300 thousand applications, including whether they were denied and the census tract of the applicant, which is a geographic region roughly corresponding to cities and towns.
We preprocess the dataset to keep records for loan applications approved or denied. Further, we use the Gazetteer files provided by the US Census Bureau\footnote{\url{https://www.census.gov/geographies/reference-files/time-series/geo/gazetteer-files.2021.html}} to associate each census tract with the coordinates of its center. In the end, we obtain a dataset, denoted as \lar, with 206,418 applications, among which 127,286 were granted (i.e., $0.62$ positive rate), geographically distributed along 50,647 locations, depicted in Figure~\ref{fig:lar}. 
We use \lar to audit spatial fairness when the measure of interest is the positive rate, i.e., in the \emph{statistical parity} sense. That is we want to investigate if all areas have the same chance of being granted a mortgage loan. The overall positive rate is $0.62$.

The second real dataset is about crime incidents in the city of Los Angeles from 2010--2019\footnote{\url{https://data.lacity.org/Public-Safety/Crime-Data-from-2010-to-2019/63jg-8b9z}}. The data contains several attributes including an approximate location of the incident. We consider the incident code, which we binarize into serious and non-serious crimes, as the label. We train a random forest classifier to predict the ``seriousness'' of the incident; the positive class is the serious crimes, while the negative the non-serious. As features, we select 7 attributes: time, code of police precinct, victim's age, sex and descent, the type of the incident location, and the weapon used. After removing entries with missing values, there are 711,852 incidents split into train/test sets with a 70\%:30\% ratio. We collect the predictions and the true labels for the incidents in the test set, and construct our dataset, denoted as \crime. The accuracy of the model is $0.78$.
We use \crime to audit spatial fairness where the measure of interest is the true positive rate. That is, in the \emph{equality of opportunity} sense, we want to investigate if the algorithm's accuracy for serious crimes is independent of location. Therefore, to apply the framework of \cite{2022_AAAI_XHJ+} and ours, we retain the predictions for the true positive labels, which results in 61,266 entries, and the overall true positive rate is $0.58$. \crime is depicted in Figure~\ref{fig:crime}.

The semi-synthetic dataset, denoted as \semisynth and depicted in Figure~\ref{fig:audit_fair}(a), contains 10,000 outcomes for locations that are randomly selected in Florida from the \lar dataset. 
The positive and negative are randomly assigned to each location with a probability of $0.5$. Hence, \semisynth is \emph{spatially fair by design}. The measure of interest is positive rate.

The synthetic dataset, denoted as \synth and depicted in Figure~\ref{fig:audit_fair}(b), contains 10,000 outcomes for locations selected uniformly at random within a rectangular area. The area is split into two halves, each containing 5,000 outcomes. However, the left half has twice as many positive outcomes as the right half does. Therefore, the positive rate in the left half is about $0.67$, while in the right half is $0.33$.

\subsection{Results on Partitionings}
\label{sec:exp:partitioning}

In the first round of experiments, we consider a setting appropriate for the \meanvar measure of spatial unfairness proposed in \cite{2022_AAAI_XHJ+}. Therefore, we consider rectangular partitionings of the space. To compare our methodology with \meanvar, we restrict our methodology to only audit for fairness the partitions that belong to the partitionings.

\stitle{Is it Fair?}
The first experiment is to test whether the two methods can correctly audit for fairness, i.e., answer the ``\emph{is it fair}'' question. For this we use the \semisynth and \synth datasets that we have designed to be fair and unfair, respectively. 
We construct 100 rectangular partitionings, where the number of horizontal and vertical splits of the space is randomly selected between 10 to 40.

As discussed in the context of Figure~\ref{fig:audit_fair}, \meanvar is $0.0522$ for the fair-by-design \semisynth and $0.0431$ for the unfair-by-design \synth; recall that lower \meanvar values suggest more fairness. Therefore, \meanvar fails to discern fairness. In contrast, our method finds that \semisynth is fair while \synth is unfair at the $0.005$ statistical significance level.

\stitle{Where is it Unfair?}
In the second set of experiments, we partition the space into a regular grid of various granularities, and investigate whether the methods can meaningfully identify partitions that potentially exhibit spatial unfairness.
For \meanvar, we identify the partitions that make the largest contribution to the \meanvar value. As discussed, in Section~\ref{sec:intro}, these are the partitions with extreme measures that differ greatly from the partitioning mean. We rank partitions according to their contribution to \meanvar.
For our method, we fix the statistical significance level at $0.005$, and identify all partitions that exhibit spatial unfairness at that significance level. We rank partitions according to their \likely value (Eq.~\ref{eq:likelihood_value}).

\begin{figure}
\centering
\begin{subfigure}[b]{\columnwidth}
\centering
\includegraphics[width=0.8\linewidth]{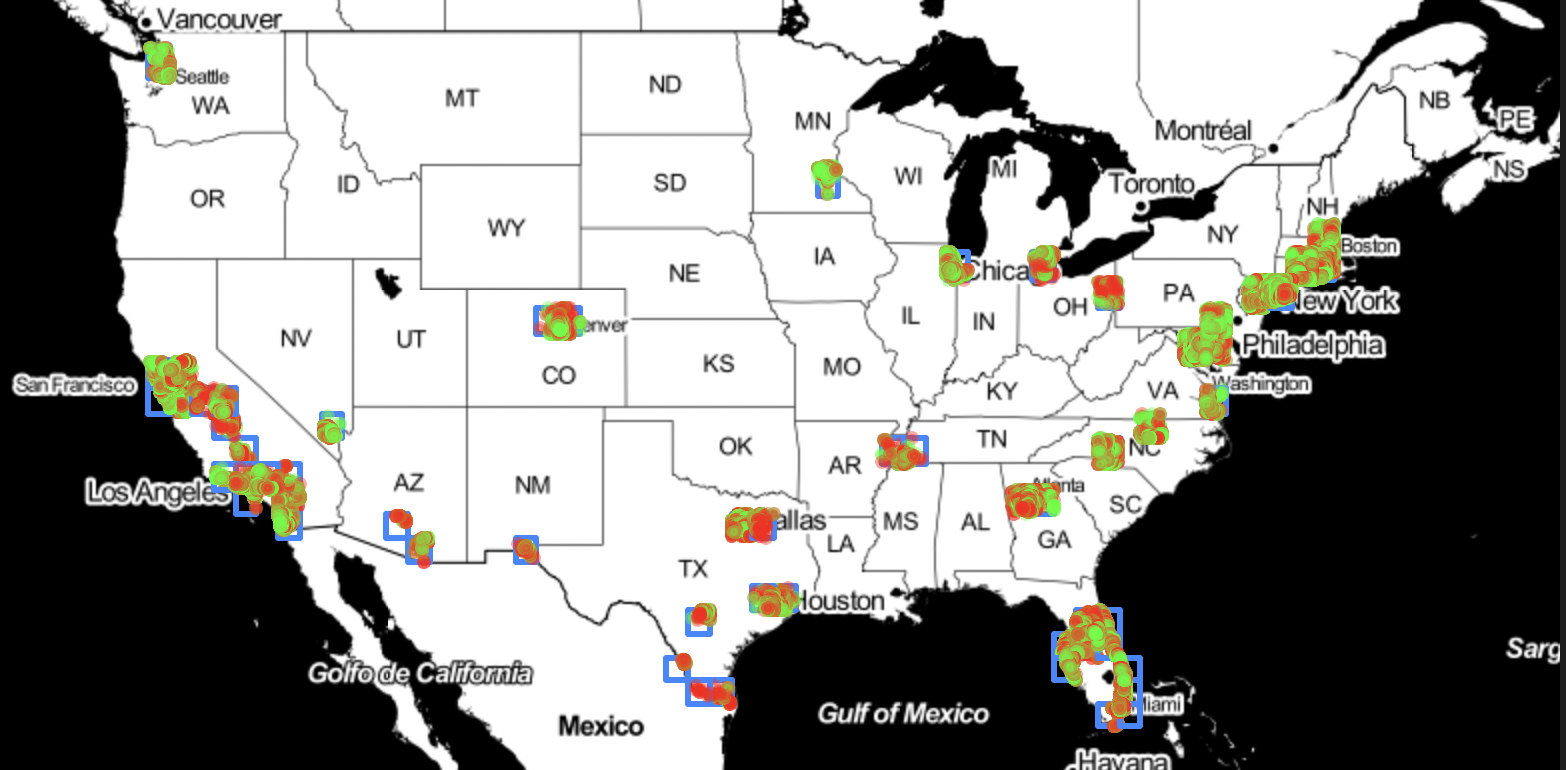}
\caption{Spatial Fairness: the 59 statistically significant unfair partitions}
\end{subfigure}\\
\begin{subfigure}[b]{\columnwidth}
\centering
\includegraphics[width=0.8\linewidth]{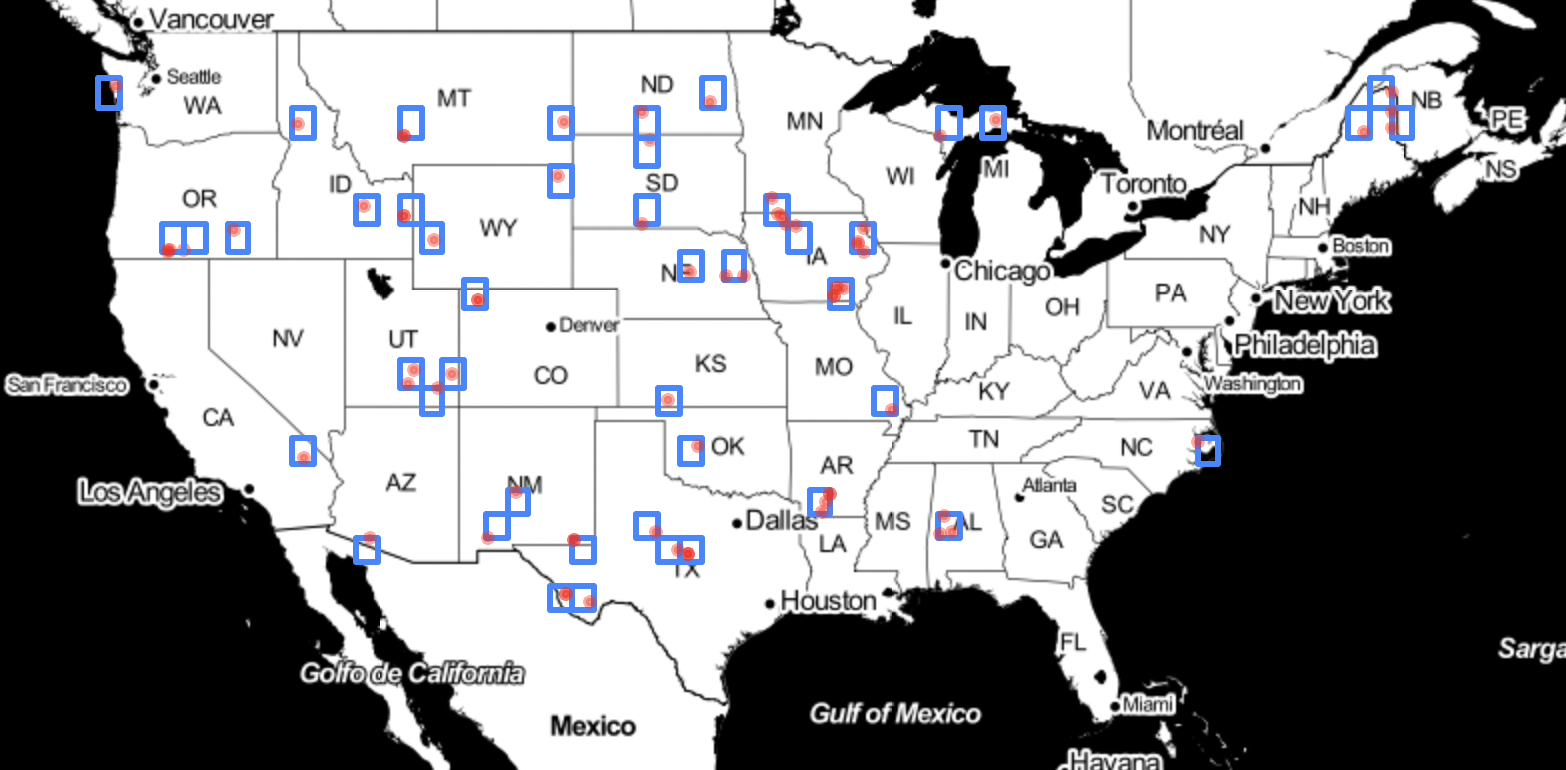}
\caption{\meanvar-Based Fairness: top-50 partitions of highest \meanvar}
\end{subfigure}
\caption{\lar: Results for a high-resolution partitioning of $100 \times 50$.}
\label{fig:grid_100x50}
\end{figure}

We first examine spatial fairness in terms of positive rate (i.e., statistical parity) and use the \lar dataset.
Intuitively, we want to identify partitions where the algorithm assigns the positive/desirable class differently from the global mean.
We consider a partitioning of high resolution $100 \times 50$.
Our framework declares the outcomes as spatially unfair and identifies 59 statistically significant partitions depicted in Figure~\ref{fig:grid_100x50}(a). The partition that has the highest \likely value is displayed in Figure~\ref{fig:pos_rates}(b), and concerns a region in northern California that covers almost 8,000 outcomes, among which 84\% are positive. Our framework identifies mostly dense regions that have statistically significant different local positive rates compared to the global.
In contrast, we display the top-50 partitions according to \meanvar in Figure~\ref{fig:grid_100x50}(b). It is easy to observe that they are all very sparse partitions that contain only negative outcomes. The largest of them with 5 outcomes is displayed in Figure~\ref{fig:pos_rates}(a).

Comparing the results of highest \meanvar and \likely in Figure~\ref{fig:pos_rates}, we find that the partition in California has a non-extreme local positive of 0.84, compared to 0 for the partition in Iowa, whereas the former has a much higher and significant log-likelihood difference of about 1000 compared to 0.96 for the latter (in this experiment log-likelihood differences beyond 9.6 are significant at the $0.005$ level).
As expected, \meanvar identifies partitions with extreme measures, whereas \likely identifies partitions that have abnormal (in the statistical likelihood sense) measures.

\begin{figure}
\centering
\begin{subfigure}[b]{\columnwidth}
\centering
\includegraphics[width=0.8\linewidth]{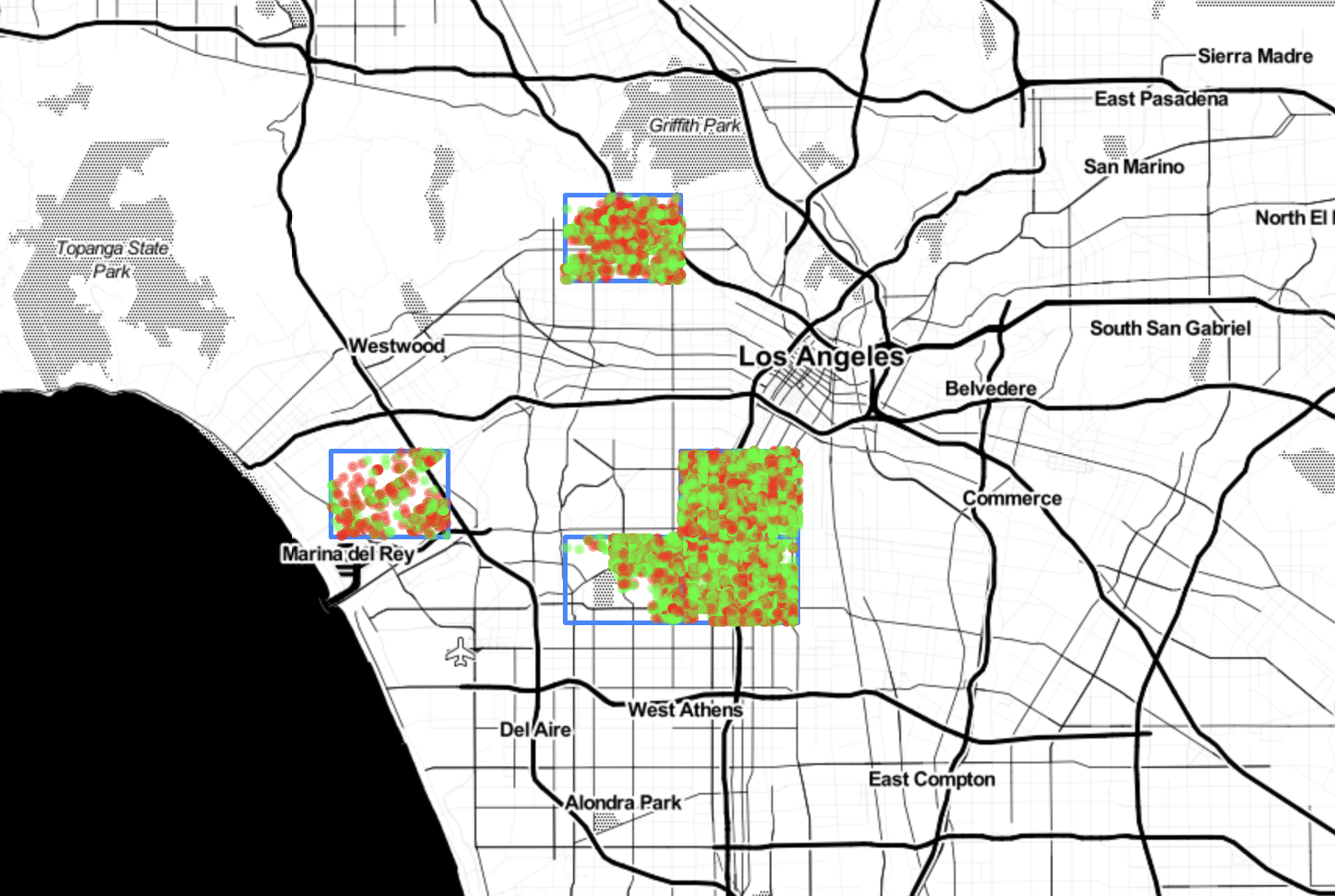}
\caption{Spatial Fairness: the 5 statistically significant unfair partitions}
\end{subfigure}\\
\begin{subfigure}[b]{\columnwidth}
\centering
\includegraphics[width=0.8\linewidth]{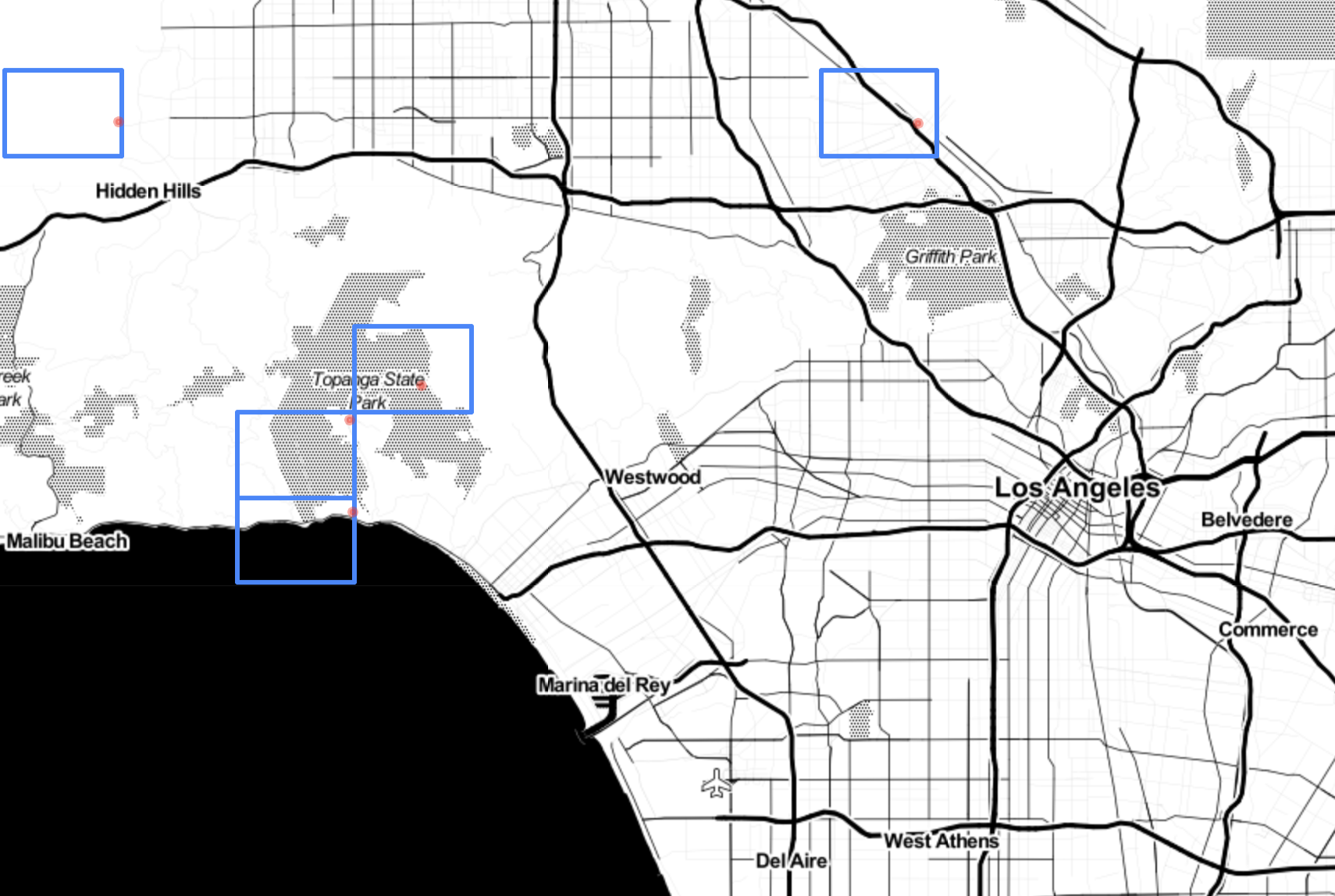}
\caption{\meanvar-Based Fairness: top-5 partitions of highest \meanvar}
\end{subfigure}
\caption{\crime: Results for a partitioning of $20 \times 20$.}
\label{fig:crime_grid_20x20}
\end{figure}

We now examine spatial fairness in terms of true positive rate (i.e., equal opportunity) and use the \crime dataset. Intuitively, we want to identify partitions where the prediction accuracy is different from the overall prediction accuracy of the algorithm.
We consider a partitioning of low resolution $20 \times 20$. 
Our framework declares the outcomes as spatially unfair and identifies 5 statistically significant partitions depicted in Figure~\ref{fig:crime_grid_20x20}(a). One of the partitions with the highest \likely value is located in Hollywood, and covers almost 3,000 outcomes, where only $51$\% of them are predicted to be serious crimes. Contrasting this to the global true positive rate of $0.58$, we see that the algorithm tends to classify incidents as non-serious in this area, more so than in other areas.
In contrast, we display the top-5 partitions according to \meanvar in Figure~\ref{fig:crime_grid_20x20}(b). All of them concern very sparse areas with a single false positive, and are thus not interesting for the auditor.

\subsection{Results for Unrestricted Regions}
\label{sec:exp:unrestricted}

In this set of experiments, we only study our framework and can thus examine arbitrarily sized regions. Specifically, we consider square regions with 20 different side lengths ranging from 0.1 up to 2 degrees (roughly 10 to 200 kilometers). The centers of square regions are placed in 100 locations defined as the centers of a k-means clustering of the observation locations. In total, we scan 2,000 square regions. Their centers and their smallest and largest shape are shown in Figure~\ref{fig:squares}.

\begin{figure}
\centering
\includegraphics[width=0.8\linewidth]{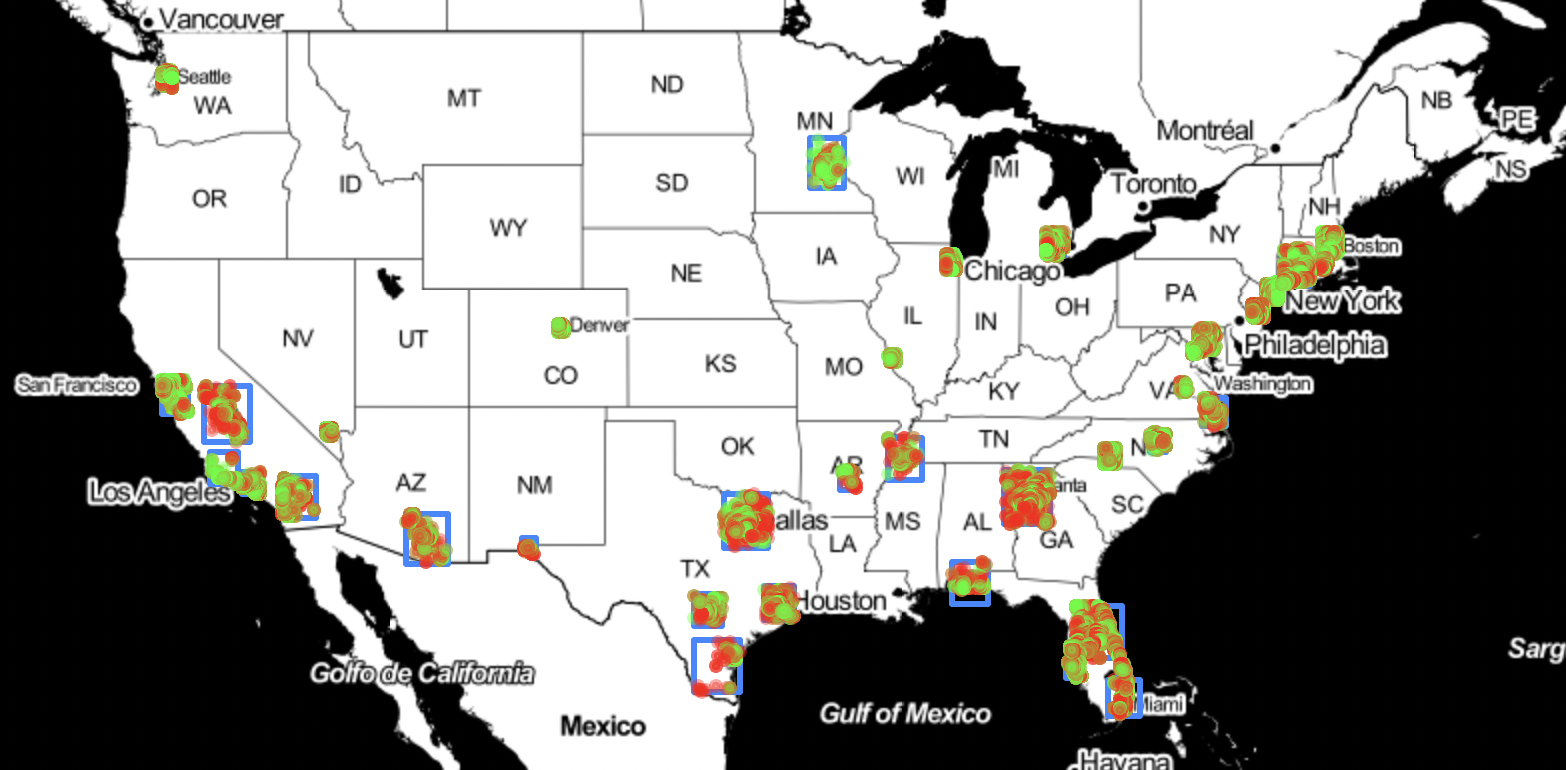}
\caption{\lar: The 28 non-overlapping unfair regions}
\label{fig:unfair}
\end{figure}

Applying our framework at a significance level of $0.005$, we identify 700 unfair regions. As these regions intersect each other, we select a set of non-overlapping regions. We examine centers in sequence, and for each center we keep the region with the highest value of the statistic. Fig~\ref{fig:unfair} displays the 28 non-overlapping regions. The important observation is that our framework identifies regions of varying area size and observation size. For example an area near Tampa, FL is the smallest radius of $0.1$ degrees with the largest number, 473, of observations, while a nearby area centered in Orlando, FL is the largest, with a radius of $1$ degrees, and 4,783 observations.

In conclusion, the regions returned as evidence from our methodology that the algorithm is unfair are non-trivial (especially, compared to those discovered by \meanvar), and can be used from the auditor to further investigate whether the observed unfairness is justified or not.

%% file: conclusion.tex
\section{Conclusion}
\label{sec:concl}

This paper introduces a generally applicable definition of spatial fairness. It includes a concrete framework to audit an algorithm for spatial fairness, and the ability to identify regions that are very likely to be spatially unfair with statistical significance. The intuition is that for any region, the distribution of outcomes inside and outside the region should be similarly distributed. The statistical test examines if the spatial fairness assumption is more likely than the alternative that allows for regions with different outcome distributions inside and outside.

\stitle{Acknowledgement:}
The authors were partially supported by the EU's Horizon Programme call, under Grant Agreements No. 101093164 (ExtremeXP) and No. 101070568 (AutoFair).

%% file: appendix.tex
\appendix

\section{Example of Spatial Fair By Design Algorithm}

\begin{figure}
\centering
\begin{subfigure}[b]{0.35\columnwidth}
\includegraphics[width=\linewidth]{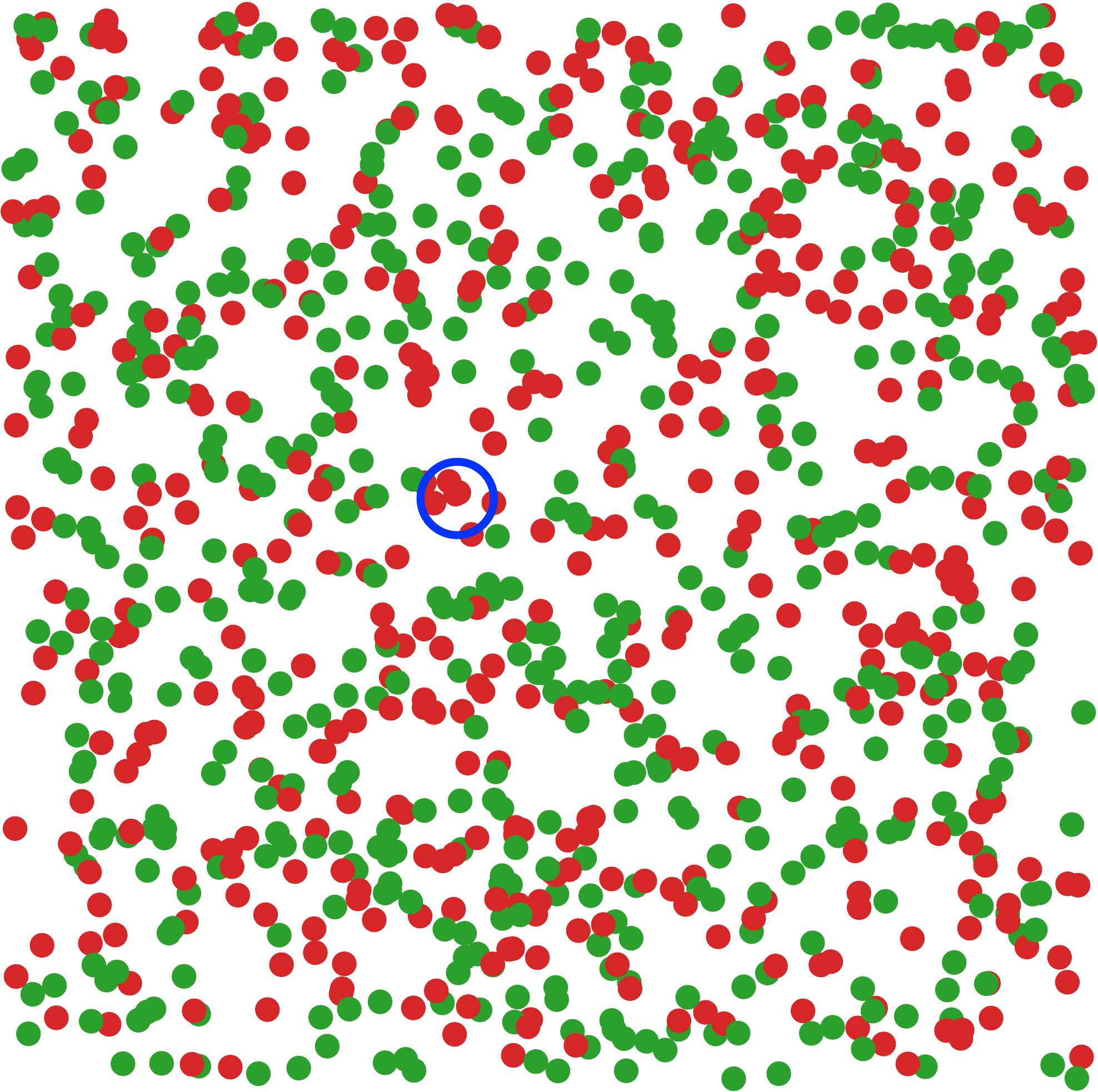}
\end{subfigure}%
\hspace{0.02\columnwidth}%
\begin{subfigure}[b]{0.35\columnwidth}
\includegraphics[width=\linewidth]{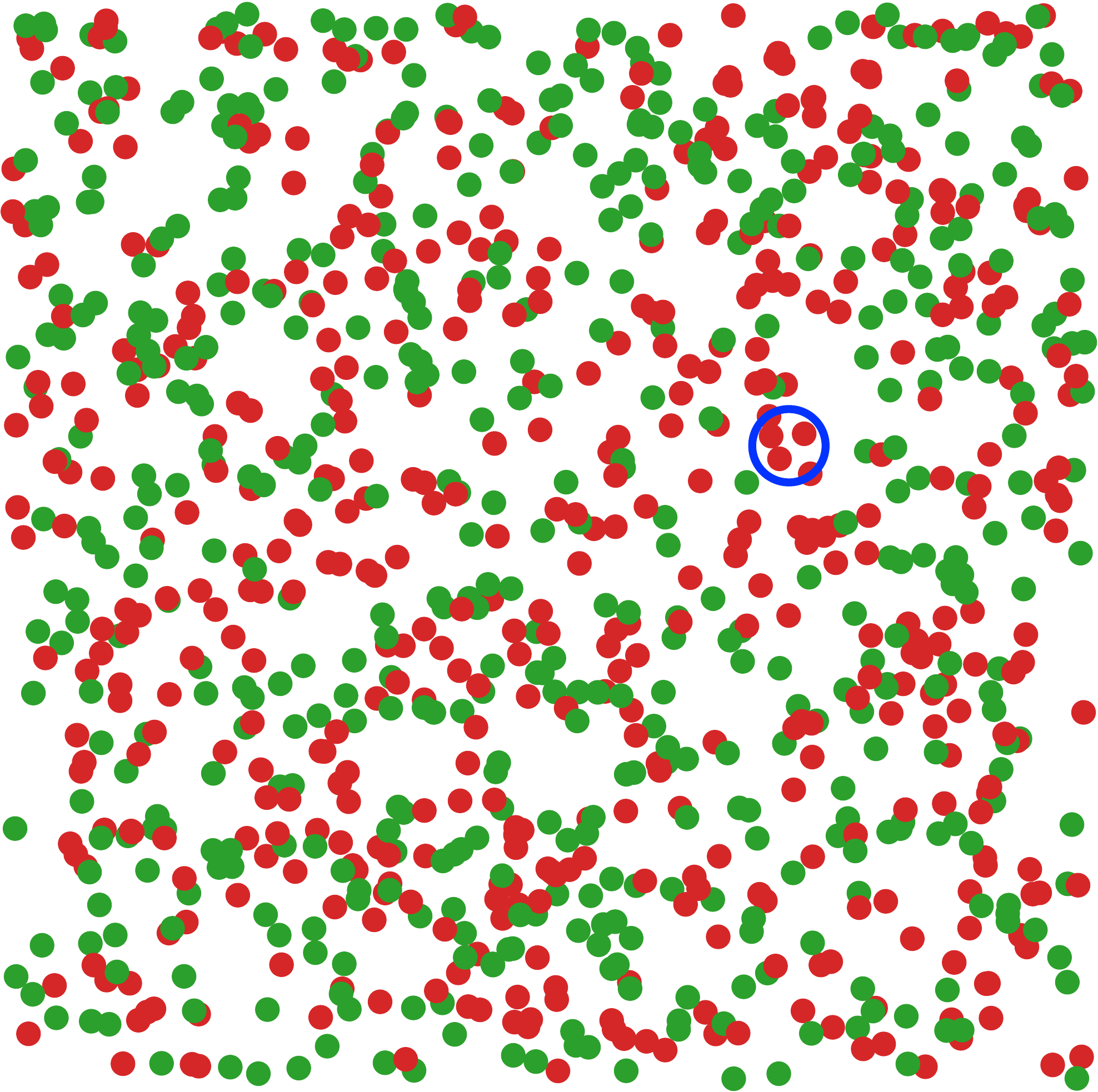}
\end{subfigure}\\ \vspace{0.02\columnwidth}
\begin{subfigure}[b]{0.35\columnwidth}
\includegraphics[width=\linewidth]{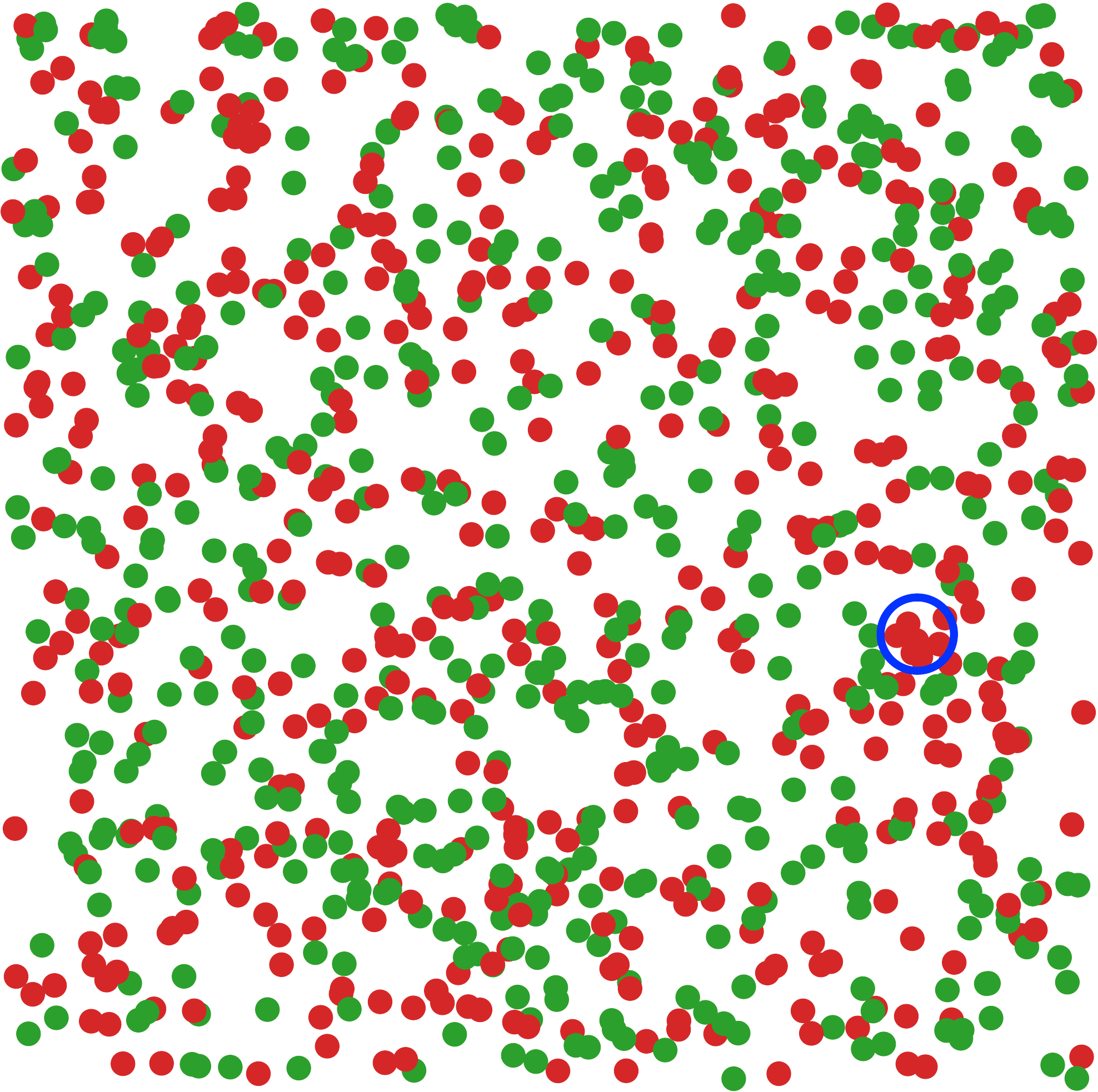}
\end{subfigure}%
\hspace{0.02\columnwidth}%
\begin{subfigure}[b]{0.35\columnwidth}
\includegraphics[width=\linewidth]{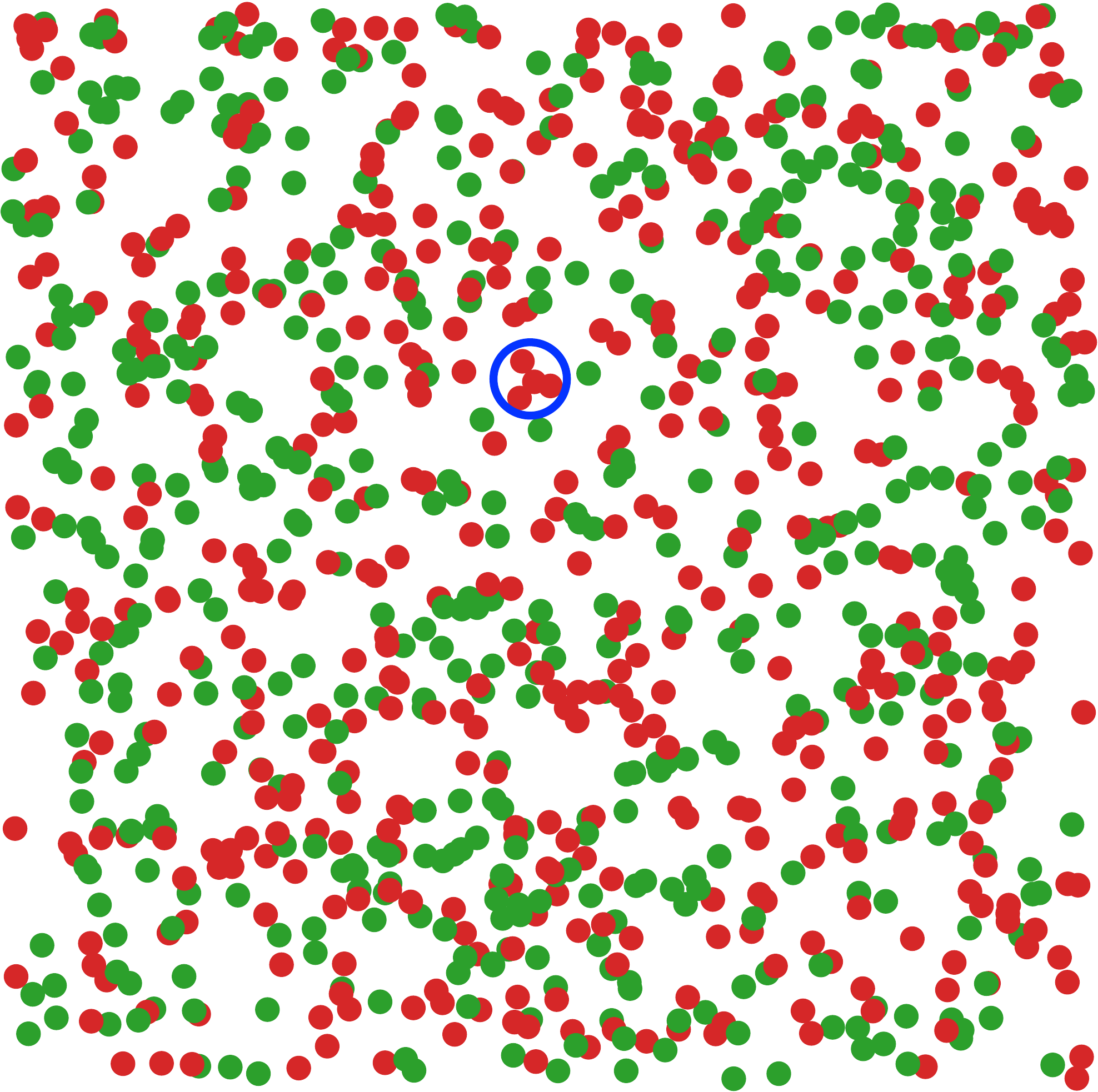}
\end{subfigure}
\caption{Four examples of 1,000 positive/negative outcomes of a spatially fair algorithm with the same positive rate 50\% and with the same spatial distribution. In all examples, it is easy to identify a region with at least five negative and no positive outcomes, drawn as a blue circle.}
\label{fig:random}
\end{figure}

Consider an algorithm that produces two outcomes, positive and negative, displayed as green and red points, respectively. For each location, there is the same probability $\rho = 0.5$ that the point is positive, independent of the location. Thus, the algorithm is spatially fair by design. 
Figure~\ref{fig:random} shows examples of possible outcomes of this spatially fair algorithm.

If we look at the regions highlighted in blue, we observe at least five negative and no positive outcomes. There is a clear discrepancy between the observed positive rate inside ($0$) and outside (roughly $\rho$) the region, and we might be tempted to claim that the algorithm is not fair. However, this event is quite likely to appear by chance. All four examples in Figure~\ref{fig:random} have the same underlying spatial distribution (locations are the same across examples) but different outcome distributions (colors change across examples). In all examples, it is easy to identify ``\emph{red}'' clusters.

\section{Additional Experiments}

We next present some additional experiments.

\subsection{Additional Results on Paritionings}

\begin{figure}
\centering
\begin{subfigure}[b]{\columnwidth}
\centering
\includegraphics[width=0.8\linewidth]{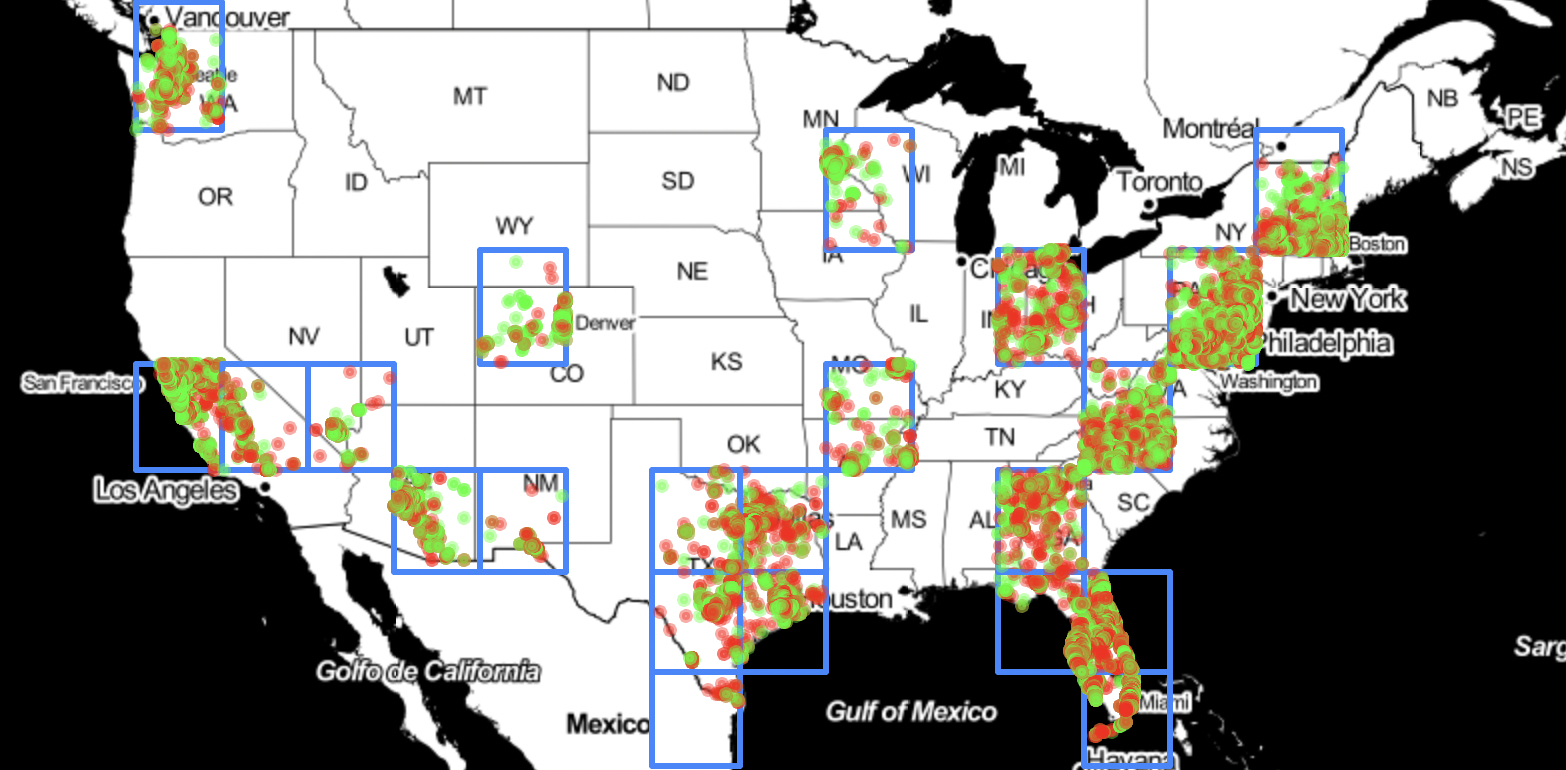}
\caption{Spatial Fairness: the 22 statistically significant unfair partitions}
\end{subfigure}\\
\begin{subfigure}[b]{\columnwidth}
\centering
\includegraphics[width=0.8\linewidth]{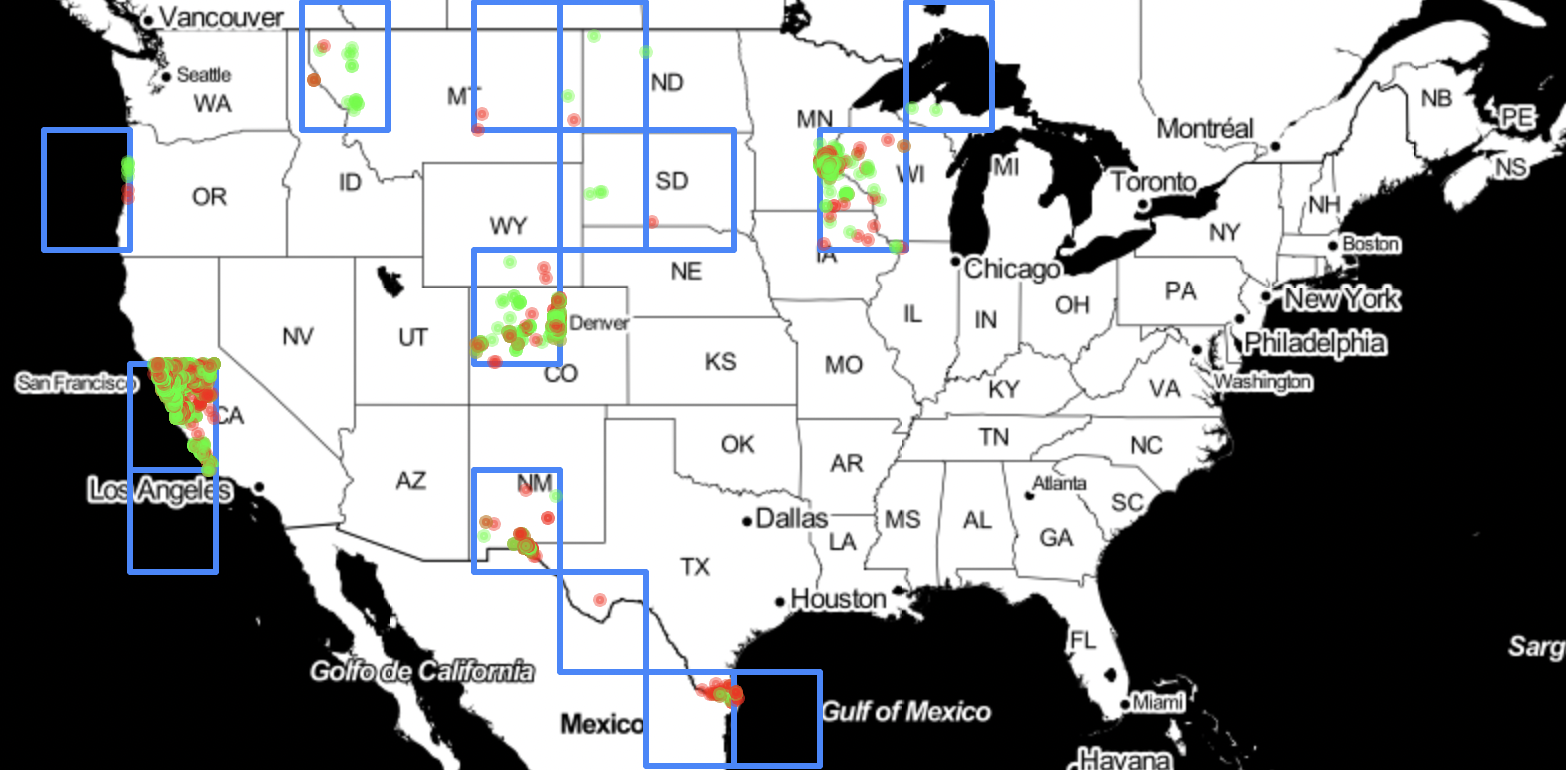}
\caption{\meanvar-Based Fairness: top-20 partitions of highest \meanvar}
\end{subfigure}
\caption{\lar: Results for a low-resolution partitioning of $25 \times 12$.}
\label{fig:grid_25x12}
\end{figure}

We consider a partitioning of low resolution $25 \times 12$. Figure~\ref{fig:grid_25x12}(a) depicts the 22 statistically significant partitions, while Figure~\ref{fig:grid_25x12}(b) shows the top-20 partitions based on the partitioning-based score. Our framework mostly identifies dense partitions as unfair, while the opposite holds for \meanvar . Nonetheless, the latter now also returns some dense areas, and also identifies the most spatially unfair region in northern California according to our definition.

\subsection{Additional Results for Unrestricted Regions}

\begin{figure}
\centering
\includegraphics[width=0.8\linewidth]{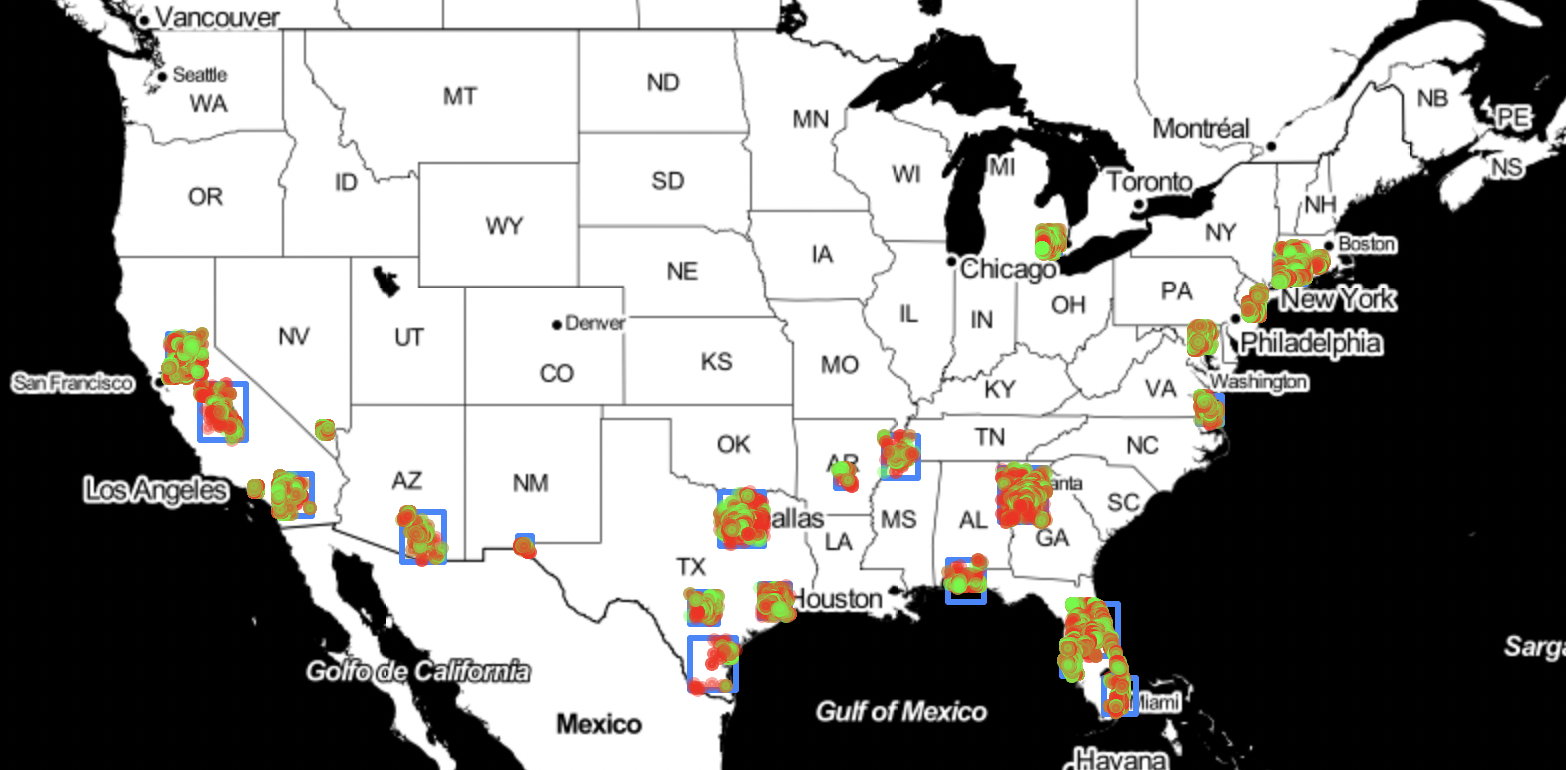}
\caption{\lar: The 27 non-overlapping unfair regions exhibiting lower positive rate inside than outside, i.e., ``\emph{red}'' regions}
\label{fig:unfair_red}
\end{figure}

We identify regions that are unfair in the sense that there are significantly less positive outcomes inside the region compared to outside. Figure~\ref{fig:unfair_red} displays the 27 non-overlapping predominantly ``\emph{red}'' regions. The most unfair red region is the one around Miami, FL with 6,281 outcomes among which only $43\%$ are positive.

\begin{figure}
\centering
\includegraphics[width=0.8\linewidth]{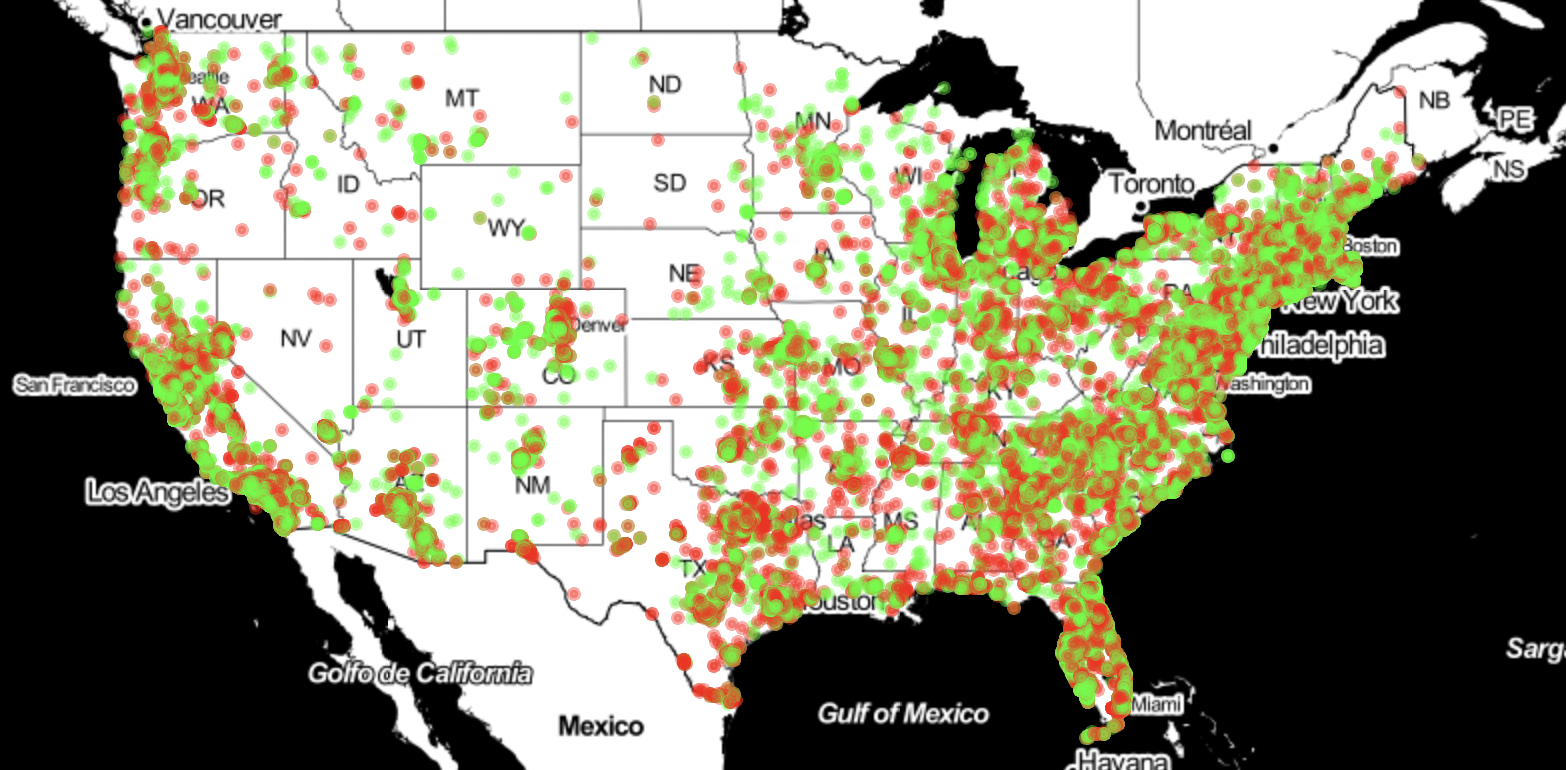}
\caption{\lar dataset depicting the locations and outcomes of mortgage loan applications; green indicates acceptance, red denial.}
\label{fig:lar}
\end{figure}

\begin{figure}
\centering
\includegraphics[width=0.8\linewidth]{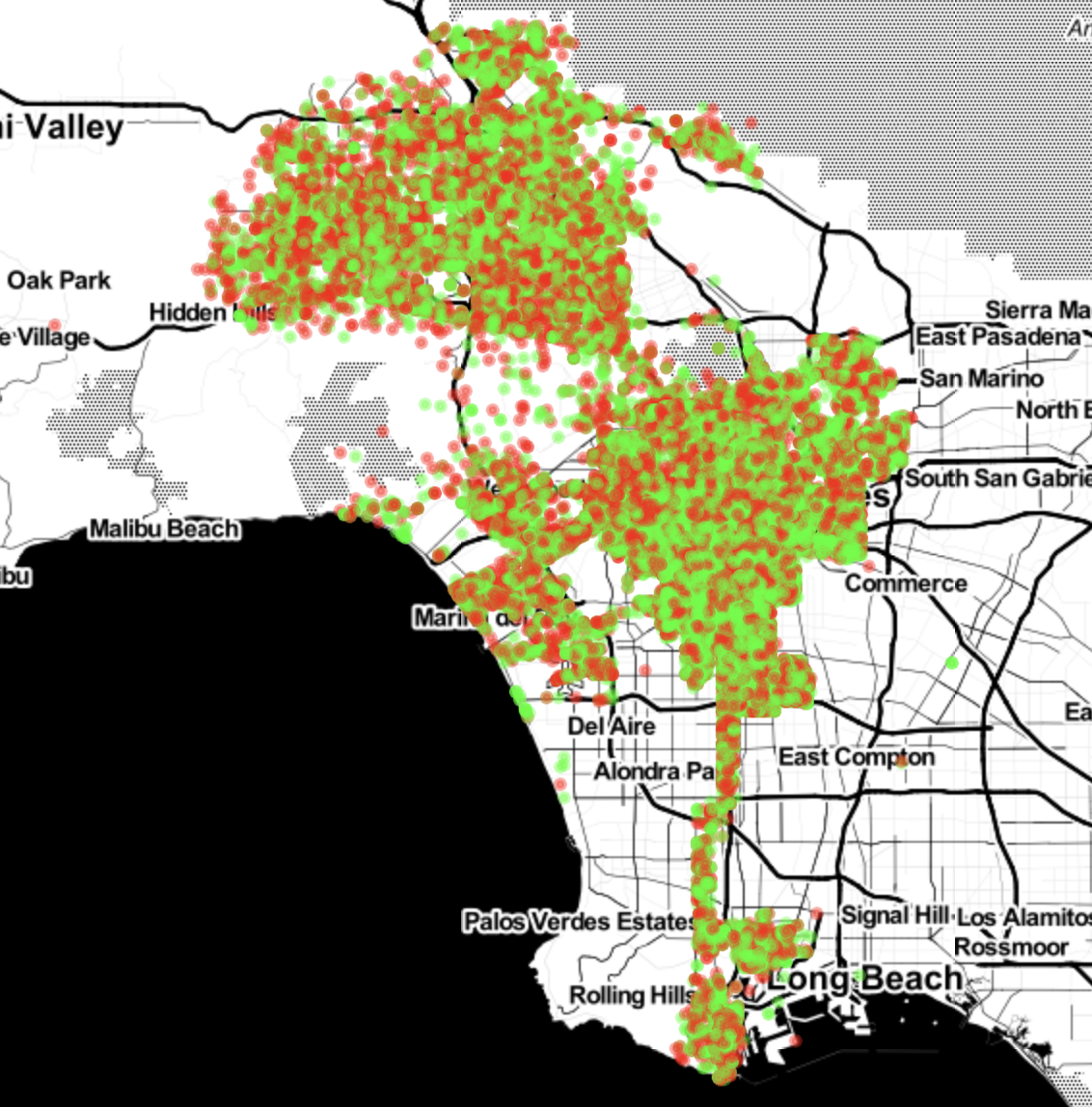}
\caption{\crime dataset depicting the locations of crime incidents; green indicates serious crimes, red non-serious crimes.}
\label{fig:crime}
\end{figure}

\begin{figure}
\centering
\includegraphics[width=0.8\linewidth]{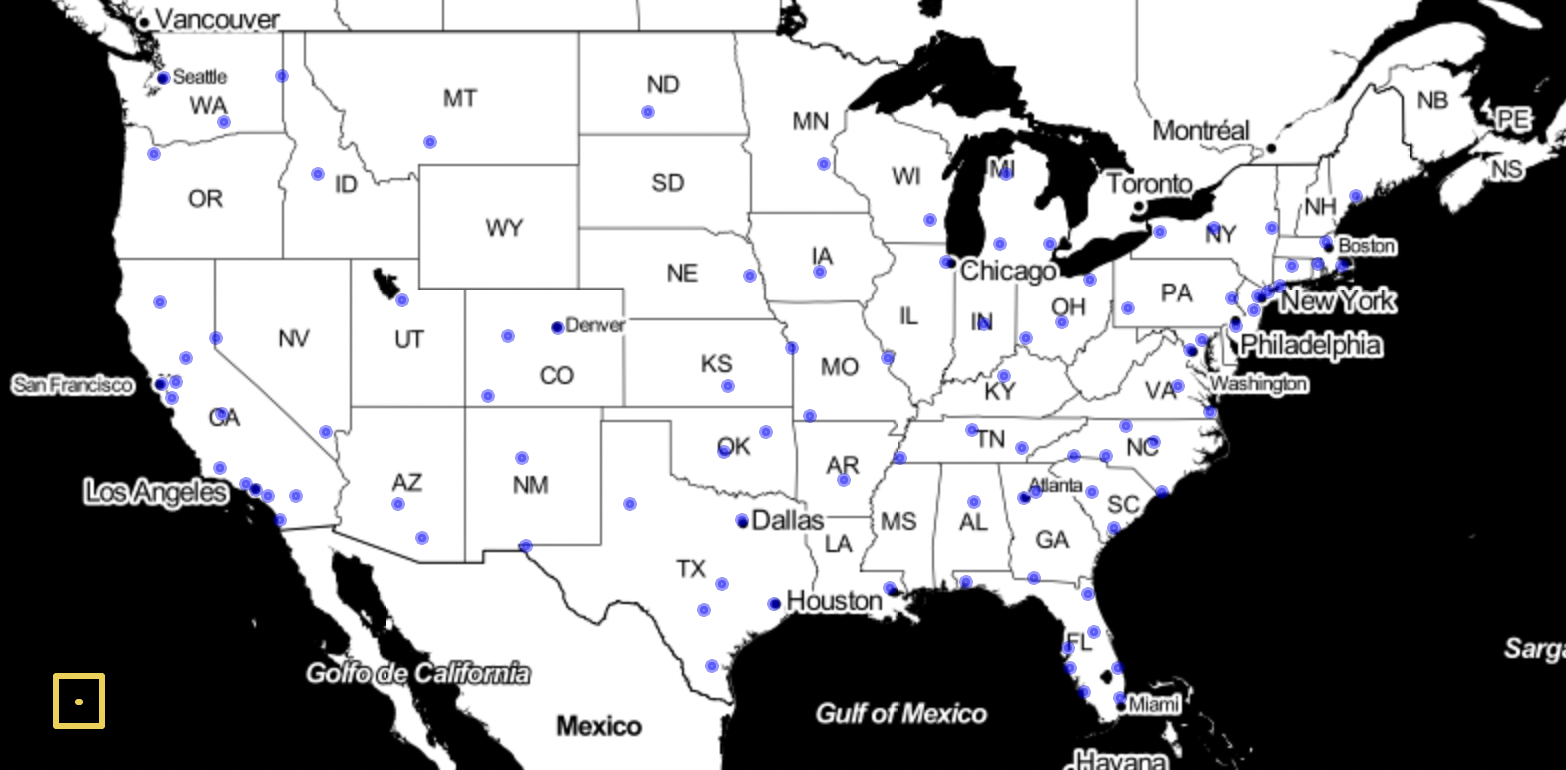}
\caption{\lar: The centers of the square regions scanned, and their smallest and largest shape}
\label{fig:squares}
\end{figure}

We look for unfair regions such that there are significantly more positive outcomes inside the region compared to outside. Figure~\ref{fig:unfair_red} displays the 27 non-overlapping ``\emph{green}'' regions. The most unfair green region is the one around San Jose, CA with 17,875 outcomes among which $83\%$ are positive.

\begin{figure}
\centering
\includegraphics[width=0.8\linewidth]{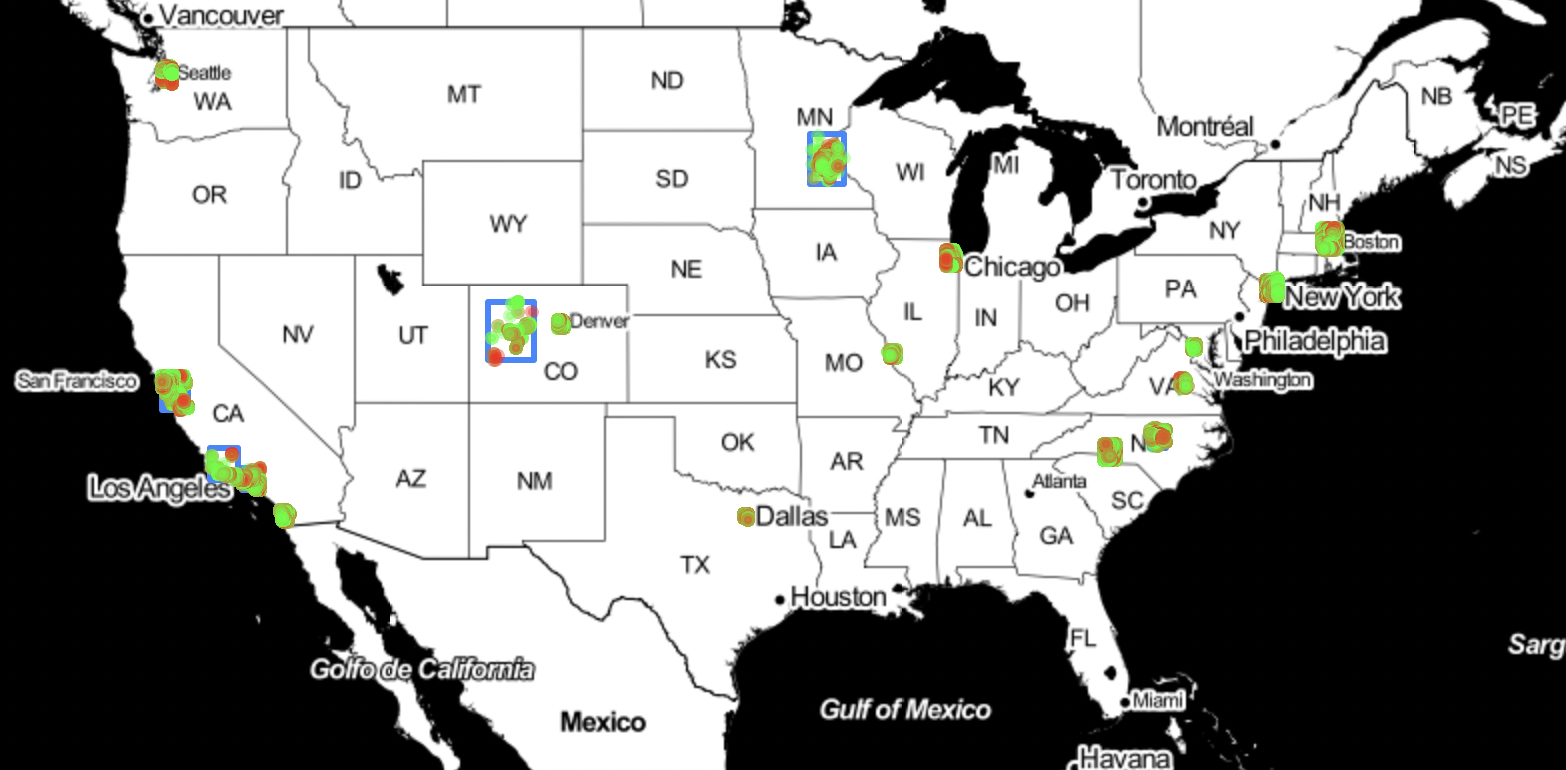}
\caption{\lar: The 17 non-overlapping unfair regions exhibiting higher positive rate inside than outside, i.e., ``\emph{green}'' regions}
\label{fig:unfair_green}
\end{figure}